\documentclass[10pt,twocolumn,letterpaper]{article}

\usepackage{iccv}
\usepackage{times}
\usepackage{epsfig}
\usepackage{graphicx}
\usepackage{amsmath}
\usepackage{amssymb}
\usepackage{xcolor}
\usepackage{multirow}
\usepackage{subcaption}
\usepackage{verbatim}
\usepackage[accsupp]{axessibility}

\usepackage{multido}
\usepackage{pdfpages}

\usepackage{pifont}
\usepackage{array}  

\usepackage{color, colortbl}

\usepackage[breaklinks=true,letterpaper=true,colorlinks,bookmarks=false]{hyperref}

\iccvfinalcopy

\newcommand{\cmark}{\ding{51}}%
\newcommand{\xmark}{\ding{55}}%

\newcommand{\notsosmall}{\fontsize{10.5pt}{12pt}\selectfont}

\newcolumntype{x}[1]{%
>{\centering\hspace{0pt}}p{#1}}

\def\iccvPaperID{1651} 

\definecolor{somegray}{rgb}{0.5, 0.5, 0.5}
\newcommand{\darkgrayed}[1]{\textcolor{somegray}{#1}}
\makeatletter
\newcommand*\titleheader[1]{\gdef\@titleheader{#1}}
\AtBeginDocument{%
  \let\st@red@title\@title
  \def\@title{%
    \vskip-3em
    \bgroup\normalfont\large\centering\@titleheader\par\egroup
    \vskip1.5em\st@red@title}
}
\makeatother

\titleheader{\darkgrayed{This paper has been accepted for publication at the \\
IEEE/CVF International Conference on Computer Vision (ICCV), Paris, 2023.
\copyright IEEE}}

\title{Active Stereo Without Pattern Projector}

\begin{document}

\author{Luca Bartolomei$^{*,\dagger}$ \hspace{0.7cm} Matteo Poggi$^\dagger$ \hspace{0.7cm} Fabio Tosi$^\dagger$ \hspace{0.7cm} Andrea Conti$^\dagger$ \hspace{0.7cm} Stefano Mattoccia$^{*,\dagger}$ \\
\notsosmall $^*$Advanced Research Center on Electronic System (ARCES) \\ 
\notsosmall $^\dagger$Department of Computer Science and Engineering (DISI) \\
\notsosmall University of Bologna, Italy \\
{\tt\small\{luca.bartolomei5, m.poggi, fabio.tosi5, andrea.conti35, stefano.mattoccia\}@unibo.it} \\
\small\url{https://vppstereo.github.io/}
}

\twocolumn[{%
\renewcommand\twocolumn[1][]{#1}%
\maketitle
\vspace{-2.5em}
\centering
   \includegraphics[trim=0cm 23cm 18cm 0cm, clip, width=0.98\textwidth]{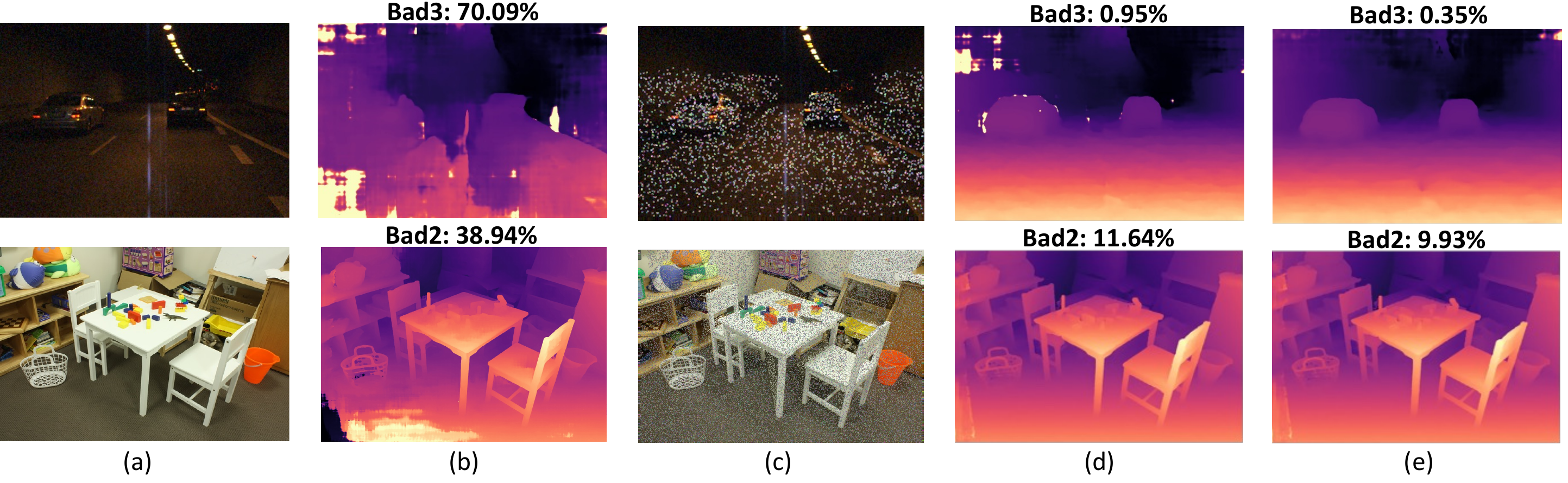}
   \vspace{-0.4cm}
   \captionof{figure}{\small\textbf{Virtual Pattern Projection for deep stereo.} 
   Either in challenging outdoor (top) or indoor (bottom) environments (a), a stereo network such as PSMNet \cite{chang2018psmnet} often struggles (b). By projecting a virtual pattern on images (c), the very same network dramatically improves its accuracy (d). Further training the model to deal with the augmented images (e) improves the results even more.}
  \label{fig:teaser}
  \vspace{0.4cm}
}]

\begin{abstract}
This paper proposes a novel framework integrating the principles of active stereo in standard passive camera systems without a physical pattern projector. We virtually project a pattern over the left and right images according to the sparse measurements obtained from a depth sensor. Any such devices can be seamlessly plugged into our framework, allowing for the deployment of a virtual active stereo setup in any possible environment, overcoming the limitation of pattern projectors, such as limited working range or environmental conditions. Experiments on indoor/outdoor datasets, featuring both long and close-range, support the seamless effectiveness of our approach, boosting the accuracy of both stereo algorithms and deep networks.
\end{abstract}

\section{Introduction}

Depth perception is crucial in several computer vision tasks, including autonomous driving, 3D reconstruction, robotics, and augmented reality. Inferring depth from standard cameras, according to different setups and strategies, is one of the most widely deployed techniques due to its low cost and potentially unbounded image resolution. At the core of these approaches, using multiple cameras or a moving one, there is the problem of determining visual correspondence. However, matching points across frames is inherently ambiguous in the presence of textureless regions, repetitive patterns, and non-Lambertian materials. This task is even more challenging when performed densely for each pixel in the input images. Although deep learning has achieved excellent results, as witnessed by the recent literature \cite{poggi2021synergies}, it is also prone to the well-known domain shift issue absent in conventional, less accurate hand-crafted methods. Specifically, since learning-based methods rely on training data, they suffer severe drops in accuracy when facing different data distributions \cite{Poggi2021continual}. 
A different approach to depth perception relies on active sensing technologies, such as LiDAR (Light Detection and Ranging), ToF (Time of Flight), and Radar (Radio Detection and Ranging). 
However, each technology has limitations. LiDAR technology is reliable but features a density much lower than the resolution of modern cameras, making it extremely expensive as density increases. ToF suffers from the same limitation and is also unreliable under sunlight and at longer distances. 
Radar allows for a more extended depth range sensing, but it is sparser, noisier, and with a narrower vertical field of view \cite{Radar_Monocular}.
Finally, active systems that estimate depth from images also exist, relying on structured \cite{Tutorial_structured_light} or unstructured \cite{INTEL_REAL_SENSE_UNSTRUCTURED} pattern projection results more accurate than passive imaging techniques and at higher resolution with respect to the aforementioned devices. However, these systems are bounded by the need for the projected pattern to be clearly visible in images, and thus cannot work beyond very close distances (i.e., a few meters), are unsuited for outdoor use with sunlight, and the presence of multiple projectors might make them interfere. 

Due to their complementary strengths and limitations, setups made of active and passive technologies are widespread for several application fields, ranging from autonomous driving, where almost all prototypes have heterogeneous sensor suites, to augmented reality with smartphones and tablets equipped with cameras and active depth sensors. Consequently, different solutions exist in the literature to exploit the synergy between active and passive depth sensing \cite{poggi2019guided,LIDARSTEREONET,wang20193d}. The common key trait of most of these sensor-fusion methods consists in modifying the internal behavior of the camera-based stereo matcher or concatenating the sparse points with the color images. 
In contrast, this paper proposes a novel paradigm to leverage the synergy between active and passive sensing. It works by coherently hallucinating the vanilla stereo pair acquired by a standard camera simplifying the visual correspondence task performed by any stereo network/algorithm as if a \textit{virtual} pattern projector were present in the scene. Such virtual coherent pattern projection is feasible by exploiting the stereo geometry and a registered active depth sensor providing sparse yet accurate measurements, like in \cite{poggi2019guided,LIDARSTEREONET,wang20193d}. Our proposal shares the same motivations of active methods based on unstructured pattern projection. However, unlike these strategies, it does not rely on a specific physical pattern projector with all the limitations outlined previously. Instead, by selecting the depth sensor that is better suited for the specific scenario, our approach can work in any environment and is agnostic to moving objects and camera ego-motion, as shown in Fig. \hyperref[fig:teaser]{1}.
Experimental results on standard stereo datasets support the following claims:

\begin{itemize}
    \item Even with meager amounts of sparse depth seeds (\eg, 1\% of the whole image), our approach outperforms by a large margin state-of-the-art sensor fusion methods based on handcrafted algorithms and deep networks.
    \item When dealing with deep networks trained on synthetic data, it dramatically improves accuracy and shows a compelling ability to tackle domain shift issues, even without additional training or fine-tuning.
    \item By neglecting a physical pattern projector, our solution works under sunlight, both indoors and outdoors, at long and close ranges with no additional processing cost for the original stereo matcher.
\end{itemize}

We believe that our proposal, dubbed \textit{Virtual Pattern Projection} (VPP), has the potential to become a standard component for depth perception and pave the way to exciting future developments in the field.

\begin{figure*}
    \centering
    \includegraphics[trim=0cm 2cm 0cm 12cm,clip,width=0.98\linewidth]{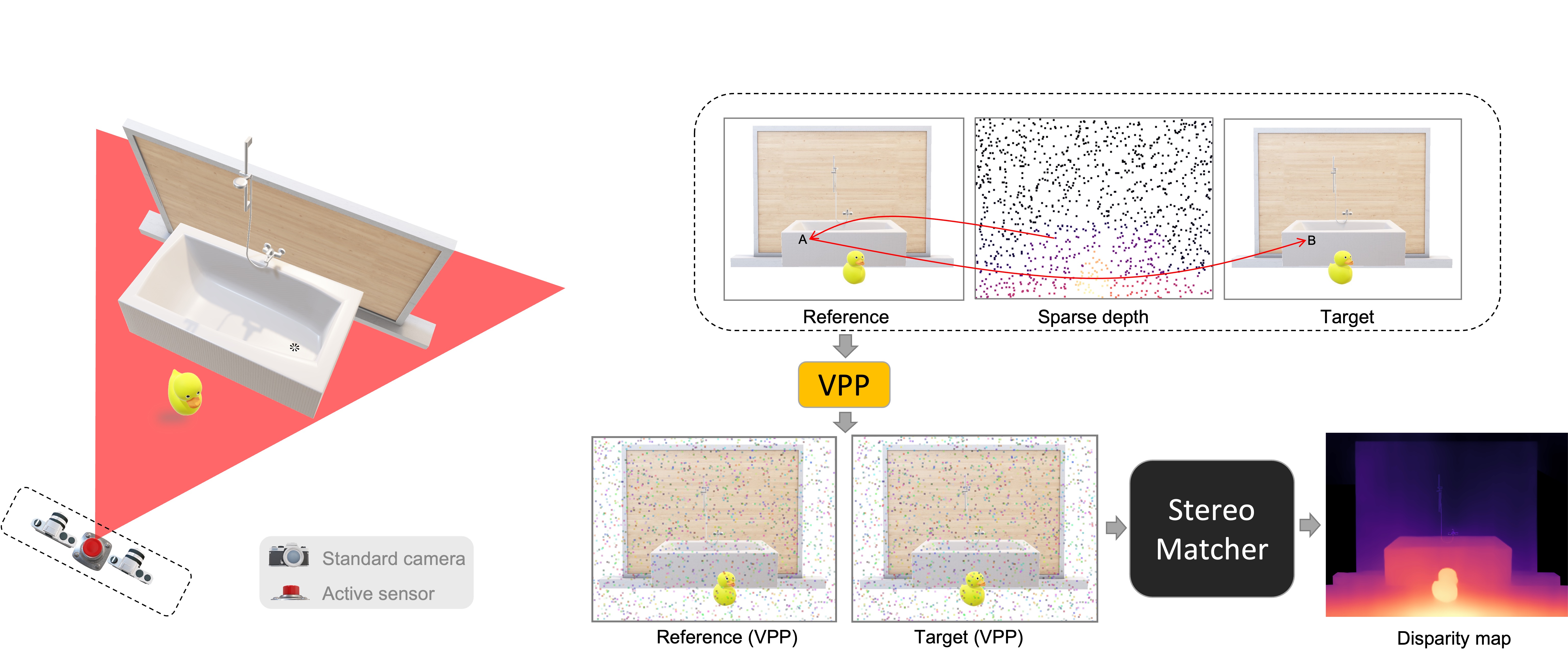}
    \vspace{-0.3cm}
    \caption{\textbf{Overview of the proposed Virtual Pattern Projection framework.} The envisioned setup on the left relies on a standard passive stereo camera and an active depth sensor. On the right, the step needed to obtain the coherently hallucinated stereo pair from the vanilla input images and sparse depth points through Virtual Pattern Projection (VPP). The patterned stereo pair in output can be processed by any stereo matcher, either handcrafted or learning-based.}
    \label{fig:VPP_method}
\end{figure*}

\section{Related Work}

\textbf{Stereo Matching.} Traditional stereo algorithms \cite{Secaucus_1994_ECCV, veksler2005stereo, yang2008stereo, yang2010constant, liang2011hardware, taniai2014graph, kolmogorov2004energy, hirschmuller2007stereo, boykov2001fast}, thoroughly investigated in \cite{scharstein2002taxonomy}, rely on handcrafted features and priors to compute dense disparity maps from stereo pairs. Deep learning has recently revolutionized stereo matching, providing significant improvements over conventional techniques on standard benchmarks \cite{zbontar2016stereo}. Specifically, end-to-end stereo networks have become the most popular and effective solution for disparity estimation. These networks can be categorized into two main families: 2D and 3D architectures, with the former adopting an encoder-decoder design \cite{mayer2016large, Pang_2017_ICCV_Workshops, Liang_2018_CVPR, saikia2019autodispnet, song2018edgestereo, yang2018segstereo, yin2019hierarchical, Tankovich_2021_CVPR}, inspired by the U-Net model \cite{ronneberger2015u}, while the latter build a feature cost volume from features extracted on the image pair and estimate the disparity map using 3D convolutions \cite{Kendall_2017_ICCV, chang2018psmnet, khamis2018stereonet, zhang2019ga, cheng2019learning, cheng2020hierarchical,  duggal2019deeppruner, yang2019hierarchical, wang2019anytime, guo2019group, Shen_2021_CVPR}, at the cost of a much higher memory requirement and runtime. A complete review of such works can be found in \cite{poggi2021synergies}. More recent works \cite{lipson2021raft, li2022practical}, instead, propose novel deep stereo networks that exploit the iterative refinement paradigm in the state-of-the-art optical flow network RAFT \cite{teed2020raft}, or rely on Vision Transformers \cite{li2021revisiting,guo2022context} to capture long-range contextual information. Despite their superior performance, deep learning-based stereo methods require amounts of annotated data for training and yet suffer from limited generalization capabilities to unseen data \cite{zhang2019domaininvariant,cai2020matchingspace,aleotti2021neural,zhang2022revisiting,liu2022graftnet,chuah2022itsa,watson2020stereo,Tosi_2023_CVPR}. To alleviate this requirement, self-supervised techniques have been proposed to train deep stereo models without ground-truth annotations by relying on photometric losses on stereo pairs or videos \cite{SsSMnet2017,Tonioni_2019_CVPR,Tonioni_2019_learn2adapt, lai2019bridging,wang2019unos,chi2021feature}, traditional algorithms and confidence measures  \cite{Tonioni_2017_ICCV, aleotti2020reversing}. Others propose self-supervised continual adaptation of stereo networks using images captured during deployment \cite{Tonioni_2019_CVPR, Poggi2021continual}.

\textbf{Active Stereo Matching.} Active stereo leverages a pattern projector, typically working in the IR spectrum, to ease the correspondence problem, particularly in low-textured and repetitive areas. These systems \cite{COMPARATIVE_ACTIVE_SENSING_SHORT_RANGE} rely on structured \cite{Tutorial_structured_light} or unstructured \cite{INTEL_REAL_SENSE_UNSTRUCTURED} light. A subclass of structured light systems -- coded light -- relies on spatial \cite{Kinect_review_PAMI} or temporal coded patterns \cite{INTEL_REAL_SENSE_STRUCTURED_PAMI}, with the latter struggling in the presence of moving objects at standard frame rate. Regardless of the technology, pattern projection is limited to shorter distances and is unsuitable for outdoor use with sunlight. 
Additionally, devices based on structured light patterns require carefully designed projectors and suffer from thermal drifts \cite{KINECT_THERMAL} among other issues \cite{KINECT_1_VS_2}. Nonetheless, active stereo is a vivid research field \cite{ACTIVE_STEREO_CVPR_2022,ACTIVE_STEREO_FANELLO}.

\textbf{Image-Guided Methods.} Another approach to estimating depth involves integrating sparse depth measurements from active sensors with RGB information. Specifically, two main trends are typically adopted in the literature.  

\textit{Depth Completion.} This approach aims to estimate dense depth maps from incomplete or noisy depth data or from a single RGB image combined with sparse depth measurements, which can be obtained from active sensors such as LiDAR or structured light. A wide range of methods have been proposed for this task, including traditional methods based on interpolation and optimization \cite{camplani2012efficient,shen2013layer,lu2014depth}, as well as deep learning-based approaches \cite{ma2019self, cheng2018depth, park2020non, hu2021penet, chen2019learning, lin2022dynamic, zhang2023completionformer}. However, these approaches typically suffer in estimating depth in regions with missing external depth values \cite{conti2023wacv}.   

\textit{Guided Stereo Matching.} In addition to advances in stereo matching, recent research has also explored the integration of stereo with active sensors. 
In Badino et al. \cite{Badino2011IntegratingLI}, the authors propose to directly integrate LiDAR data into the stereo algorithm using dynamic programming. Gandhi et al. \cite{gandhi2012high}, instead, propose a method to fuse time-of-flight (ToF) camera and stereo pairs using an efficient seed-growing algorithm. More recent works, instead, exploit depth measurements, either by concatenating them as input of CNN-based architectures \cite{LIDARSTEREONET, park2018high, zhang2020listereo, wang20193d} or by using them to guide the cost aggregation of existing cost volumes \cite{poggi2019guided, huang2021s3, zhang2022lidar, wang20193d}.
Our approach augments the given stereo pair to enhance RGB images, providing more discriminative information to the network and making it easier to solve the correspondence problem, unlike other methods that only concatenate sparse depths or guide cost aggregation.

\section{Virtual Pattern Projection (VPP)}

This section describes our approach to ease visual correspondence across images acquired by a passive stereo camera by means of the coherent virtual projection of patterns leveraging the availability of reliable sparse depth points of the same scene. 
The motivation for enriching the visual image content of the original scene is the same as methods relying on a physical projector: it aims at increasing local distinctiveness to make the visual correspondence task more robust. However, in contrast to previous methods, our strategy follows an entirely different path discarding the need for a physical projector with all its mentioned limitations. 

\subsection{Virtual Projection Principle}

Our proposal is based on the observation that given a rectified stereo rig, and availability of a set of sparse depth points registered with the reference image, each of these points implicitly allows us to determine the corresponding pixels in the stereo pair. Hence, by knowing this correspondence, we can augment the visual appearance of both pixels in the two images to make them as similar as possible and as distinctive as possible from their neighbors, as if an ideal smart virtual projector were sending its signal onto the sensed scene. 
Fig. \ref{fig:VPP_method} outlines our basic virtual projection principle, showing on the left the envisioned setup consisting of a calibrated stereo camera and a sensor performing sparse yet accurate depth estimation, registered with the stereo reference camera through an initial calibration. The vanilla stereo pair acquired by the passive cameras and the input depth map are shown at the top. Since the stereo rig and the active sensor are calibrated, the 
depth $z(x,y)$ of each sparse point allows us to locate corresponding points $I_{L}(x,y)$ and $I_{R}(x',y)$ in the two input images. For this purpose, we transform $z(x,y)$ into a disparity $d(x,y)$ by knowing \cite{SZELISKI_BOOK} the focal length $f$ 
and the baseline $b$ of the stereo camera with $d(x,y) = \frac{b \cdot f}{z(x,y)}$. Such a disparity value represents the offset needed to obtain the location along the same epipolar line of the corresponding point $I_{R}(x',y)$ in the target image with $x' = x - d(x,y)$. Fig. \ref{fig:VPP_method} shows how the two corresponding points A e B can be determined by knowing the depth of A in the input sparse depth map, and then we can augment them to achieve our goal.

\begin{figure}[t]
    \centering
    \includegraphics[trim=0cm 5cm 0cm 8cm, clip, width=1.0\linewidth]{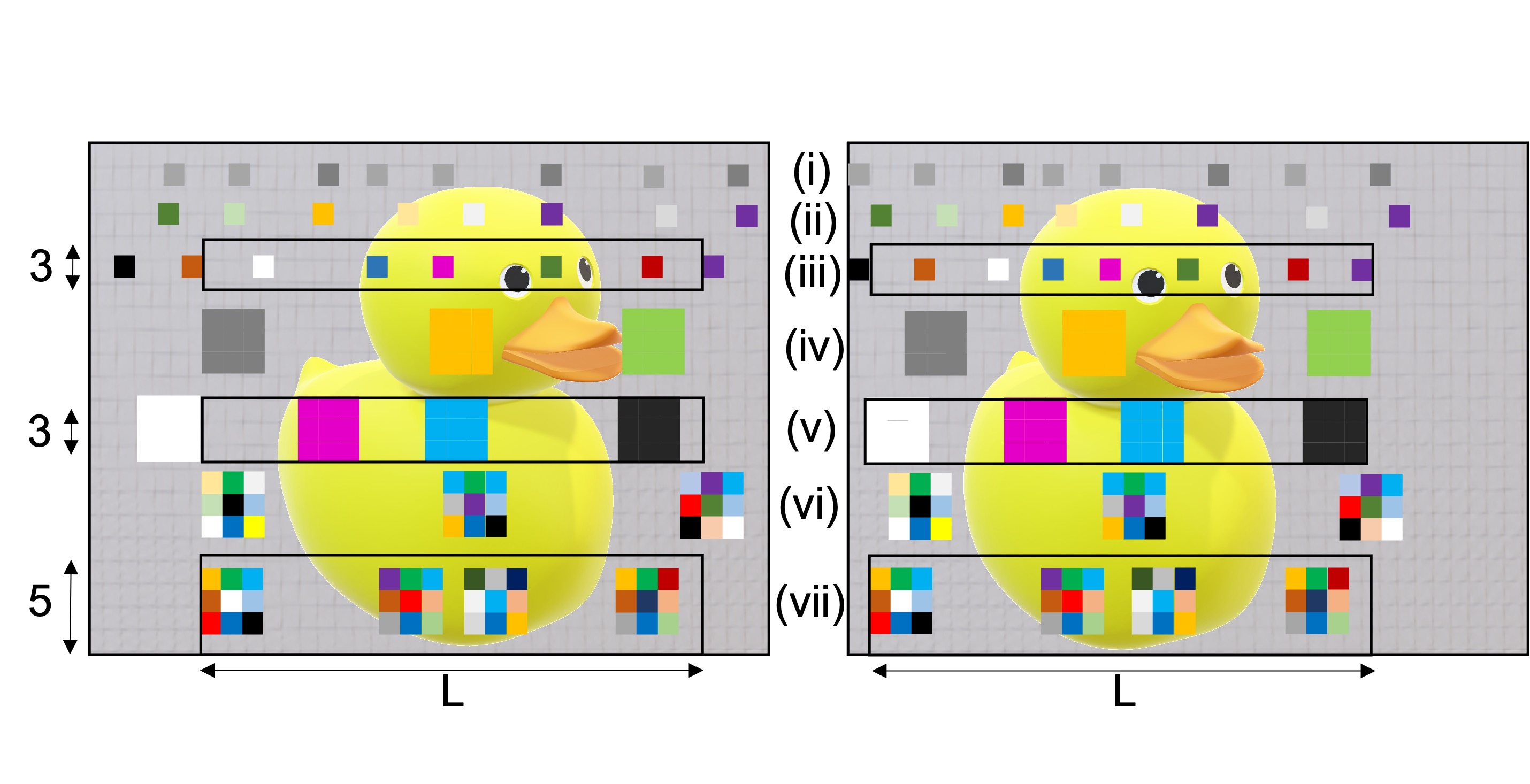}
    \vspace{-0.3cm}
    \caption{\textbf{Virtual patterns.} 
    From top to bottom: (i) indistinctive, (ii) randomly generated, (iii) distinctive, (iv) randomly generated  patch-based uniform, (v) distinctive patch-based uniform, (vi) randomly generated patch-based, and (vii) distinctive patch-based patterns. For (iii), (v), and (vii) we report the search area of length $L$, set to 64 in our experiments, and height as depicted for $3 \times 3$ virtual patterns.}
    \label{fig:Duck_pattern}
\end{figure}

\subsection{Virtual Patterns}

To properly hallucinate images and ease matching, corresponding points should be as similar as possible. Accordingly, we propose two augmenting strategies: 
\textit{random pattern} and \textit{histogram-based pattern}. Both operate on corresponding points $(x,y)$ and $(x',y)$ located along the epipolar line 
with the same pattern $\mathcal{A}(x,x',y)$ in the two images:  

\begin{equation}
    \begin{split}
        I_L(x,y) \leftarrow \mathcal{A}(x,x',y)\\
        I_R(x',y) \leftarrow \mathcal{A}(x,x',y)
    \end{split} 
    \label{eq:stereo_pattern}
\end{equation}

Their difference consists in how the operator $\mathcal{A}(x,x',y)$ generates the virtual pattern superimposed on the input images. 
Fig. \ref{fig:Duck_pattern} outlines the several possible patterns that we will discuss in the remainder.
Furthermore, as $x'$ is unlikely to be an integer, $\mathcal{A}(x,x',y)$ will apply to both $I_R(\lfloor x'\rfloor,y)$ and $I_R(\lceil x'\rceil,y)$ by means of a weighted splatting on the two.

\begin{equation}
    \begin{split}
        I_R(\lfloor x'\rfloor,y) \leftarrow \beta I_R(\lfloor x'\rfloor,y) + (1-\beta) \mathcal{A}(x,x',y) \\
        I_R(\lceil x'\rceil,y) \leftarrow (1-\beta) I_R(\lceil x'\rceil,y) + \beta \mathcal{A}(x,x',y)
    \end{split}
    \label{eq:stereo_pattern_splatting}
\end{equation}
where $\beta = x'-\lfloor x'\rfloor$ is the splatting weight.

\subsubsection{Random Patterning}

To ease visual correspondence, the pattern should both increase similarity across images, as well as distinctiveness \cite{Distinctiveness_ICCV_07} along the epipolar line. As such, 
a pattern as the one shown in Fig. \ref{fig:Duck_pattern} (i) would be suboptimal.
To address this issue, a method uses an operator $\mathcal{A}(x,x',y)$ that samples a random value from a uniform distribution 

\begin{equation}
    \mathcal{A}(x,x',y)\sim\mathcal{U}(0,255)
    \label{eq:method_3}
\end{equation}
Fig. \ref{fig:Duck_pattern} (ii) depicts a possible outcome of this approach whose extension to color images, as for any method discussed, is straightforward by applying the same strategy on each color channel. 
This operator is fast and introduces distinctiveness to some degree, although not entirely. On the contrary, an operator taking into account the image content itself could enforce stronger distinctiveness \cite{Distinctiveness_ICCV_07}.

\subsubsection{Histogram-based Patterning}

We argue that the patterns superimposed onto the vanilla image should stand out from the background and be unambiguous (at least) within nearby pixels along the same horizontal scanline, as depicted in Fig. \ref{fig:Duck_pattern} (iii).

\begin{figure}
    \centering
    \includegraphics[trim=0cm 3cm 0cm 8cm, clip, width=1.0\linewidth]{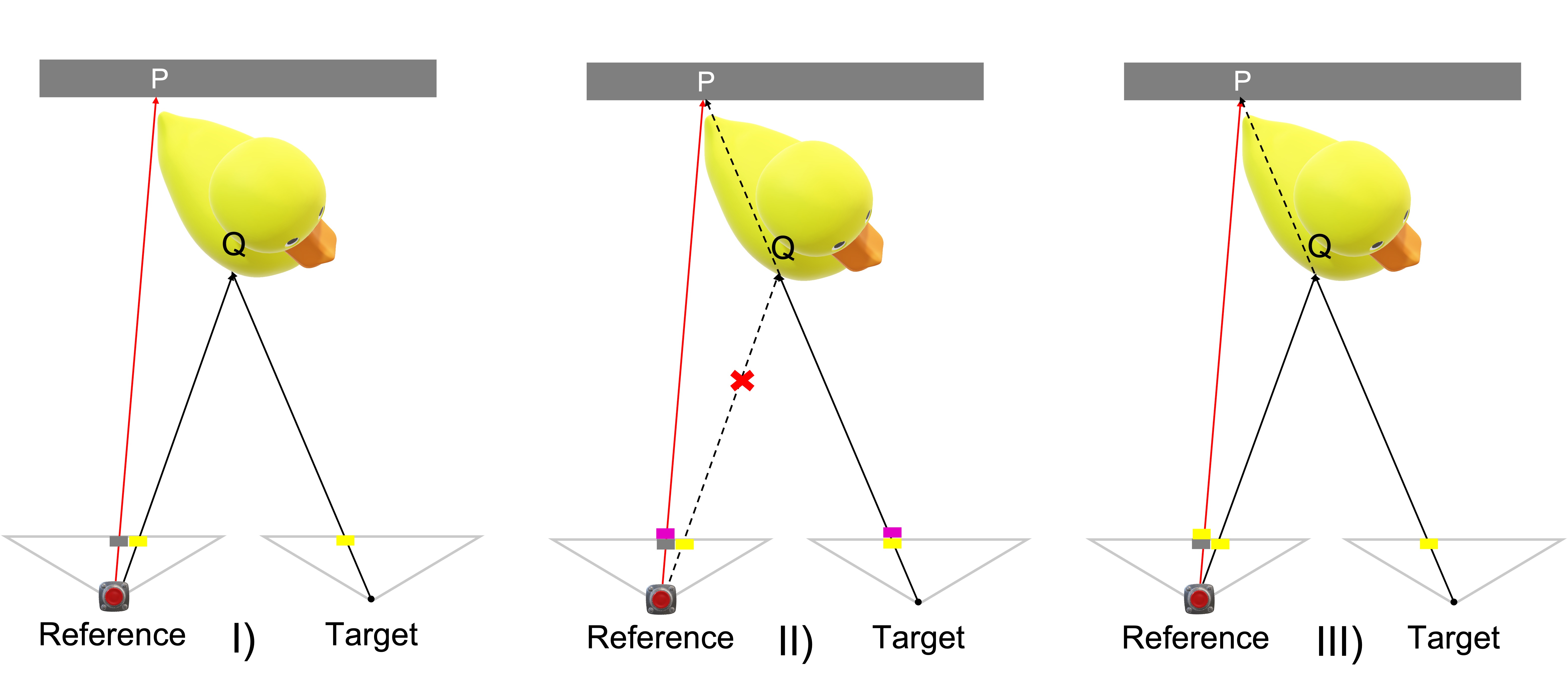}
    \vspace{-0.3cm}
    \caption{\textbf{Handling occlusions.} I) P is framed by the reference camera and depth sensor but occluded in the target camera (``NO" projection), II) Projection of the same pattern (violet) onto the two input images according to depth in the background (``BKGD"), III) Projection of the foreground image content (Q) from the target image to the background in the reference image (``FGD").}
    \label{fig:Duck_occlusion}
\end{figure}

\begin{figure*}
    \centering
    \includegraphics[trim=0cm 0cm 0cm 0.7cm,clip,width=0.98\linewidth]{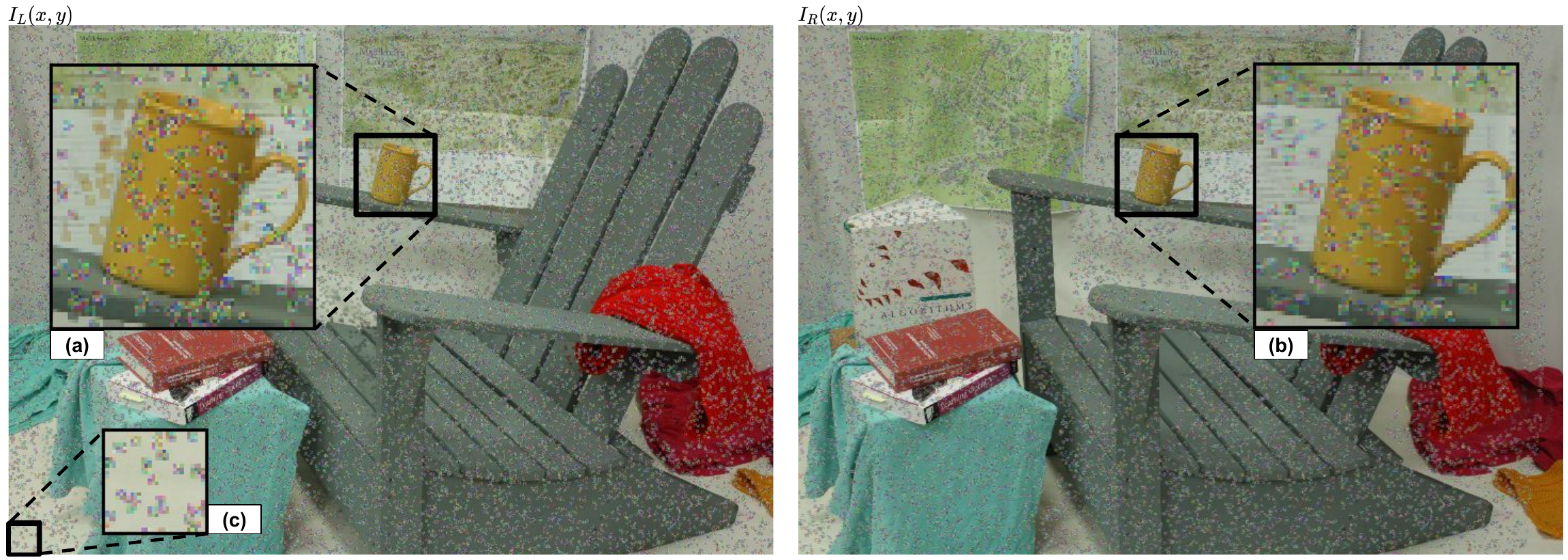}
    \caption{\textbf{VPP in action.} Hallucinated left and right images, zoomed-in view of the type (vi) pattern, showing FGD-Projection (a), corresponding area (b) with sub-pixel splatting and left border occlusion projection (c).}
    \label{fig:pattern_qualitative1}
\end{figure*}

To achieve this goal, we employ a histogram-based operator $\mathcal{A}(x,x',y)$, to select the pattern by analyzing the scanline content in the two hallucinated images. For $(x,y)$ in the reference image, we consider two windows of height 3 and length $L$ centered on it and on $(x',y)$ in the target image. Then, the histograms computed over the two windows 
are summed up and the operator $\mathcal{A}(x,x',y)$ picks the color maximizing the distance from any other color in the histogram $\text{hdist}(i)$, with $\text{hdist}(i)$ returning the minimum distance from a filled bin in the sum histogram $\mathcal{H}$

\begin{equation}
    \begin{split}
        \text{hdist}(i) = & \big\{\min\{ |i-i_l|,\,|i-i_r| \},\\
        & i_l\in[0,i[:\mathcal{H}(i_l)>0,\\
        & i_r\in]i,255]:\mathcal{H}(i_r)>0 \big\}
    \end{split}
    \label{eq:method_2_1}
\end{equation}
If every bin in $\mathcal{H}$ is filled, the color with minimum occurrence is selected.

\subsection{Advanced Virtual Patterns}

We now extend the strategies described so far, taking into account locality, the original image content, and occlusions.
\subsubsection{Locality}

The pointwise patterning strategy can be applied to larger areas to enhance the visual appearance further. To this purpose, we can extend previous methods to patches (\eg, $3 \times 3$ or $5 \times 5$), implicitly assuming the same disparity within these small regions. 
Fig. \ref{fig:Duck_pattern}, in (iv) and (v), shows the outcome of the operator $\mathcal{A}(x,x',y)$ that selects a uniform color within the patches according to, respectively, random sampling or a selection based on histograms in a larger region $L$. 
Further patterns can be generated by (vi) random sampling or (vii) histogram-base selection being performed for every pixel in the patch independently.

\subsubsection{Alpha-Blending}

Although a virtual pattern eases the match for traditional algorithms \cite{hirschmuller2007stereo}, it might hinder a deep stereo model not used to deal with it.
Thus, we combine the original image content with the virtual pattern through alpha-blending \cite{SZELISKI_BOOK} as follows, being $\alpha$ a hyperparameter:    

\begin{equation}
    \begin{split}
        I_L(x,y) \leftarrow (1-\alpha) I_L(x,y) + \alpha \mathcal{A}(x,x',y)\\
        I_R(x',y) \leftarrow (1-\alpha) I_R(x',y) + \alpha \mathcal{A}(x,x',y)
    \end{split}
    \label{eq:stereo_pattern_blending}
\end{equation}

\begin{table*}
\centering
\renewcommand{\tabcolsep}{18pt}
\scalebox{0.8}{
\begin{tabular}{|c|cccc|ccc|}
\hline
VPP & \multicolumn{4}{c|}{Hyperparameters} & \multicolumn{3}{c|}{Error Rate (\%) $>2$} \\
\hline
& Pattern & $\alpha$ & Patch & Occ. & RAFT-Stereo \cite{lipson2021raft} & PSMNet \cite{chang2018psmnet} & rSGM \\
\hline\hline
\xmark & \xmark & \xmark & \xmark & \xmark & 11.5 & 29.3 & 34.3\\
\hline

\cmark & (ii) & \xmark & \xmark & BKGD & 5.2 & 15.3 & 20.6 \\
\cmark & (iii) & \xmark & \xmark & BKGD & 5.1 & 15.2 & 20.2 \\

\cmark & (ii) & 0.4 & \xmark & BKGD & 5.8 & 16.7 & 21.2 \\
\cmark & (iii) & 0.4 & \xmark & BKGD & 5.6 & 16.1 & 20.5 \\

\cmark & (iv) & 0.4 & $3\times3$ & BKGD & 4.9 & 15.0 & 16.7 \\
\cmark & (v) & 0.4 & $3\times3$ & BKGD & 5.0 & 15.1 & 16.2 \\

\cmark & (vi) & 0.4 & $3\times3$ & BKGD & 5.0 & 15.3 & 15.9 \\
\cmark & (vii) & 0.4 & $3\times3$ & BKGD & 5.0 & 15.2 & 15.9 \\

\cmark & (iv) & 0.4 & $3\times3$ & NO & 4.9 & 14.7 & 16.2 \\
\cmark & (v) & 0.4 & $3\times3$ & NO & 4.9 & 14.8 & 15.7 \\

\cmark & (vi) & 0.4 & $3\times3$ & NO & 5.1 & 14.9 & 15.7 \\
\cmark & (vii) & 0.4 & $3\times3$ & NO & 4.9 & 14.9 & 15.4 \\

\cmark & (iv) & 0.4 & $3\times3$ & FGD & \textbf{4.8} & \textbf{14.4} & 16.1 \\
\cmark & (v) & 0.4 & $3\times3$ & FGD & \textbf{4.8} & \textbf{14.4} & 15.6 \\
\rowcolor{yellow}
\cmark & (vi) & 0.4 & $3\times3$ & FGD & 5.0 & 14.6 & 15.6 \\
\cmark & (vii) & 0.4 & $3\times3$ & FGD & \textbf{4.8} & \textbf{14.4} & \textbf{15.3} \\
\hline
\end{tabular}}
\vspace{-0.3cm}
\caption{
\textbf{Ablation on main projection hyperparameters.}
Results on Midd-A. Networks trained on synthetic data. 
}
\label{tab:ablation}
\end{table*}

\subsubsection{Occlusions}

Since occluded regions inevitably exist in a stereo setup, even assuming a depth sensor perfectly aligned with the reference camera (although this is not always the case \cite{Conti_confidence_IROS_2022}), we might not be able to project the pattern consistently on the two views, as depicted in Fig. \ref{fig:Duck_occlusion} I).
Accordingly, it is crucial to detect points hitting occluded regions to avoid projecting the same pattern on both the occluded and the occluder pixels, respectively, on the reference and target images.
Purposely, we devise a simple yet effective heuristic to 
classify sparse disparities warped onto the target image, according to the difference in disparity and spatial distance from other sparse points inferred by the sensor.

Specifically, we warp disparity $d$ for $(x,y)$ into an image-like grid $W$ at coordinates $(x',y)$. In case of collisions  -- i.e., multiple $d$ warped at the same location $(x',y)$ -- the largest $d$ is kept.
Then, each $(x_o,y_o)$ in $W$ is classified as occluded if the following inequality holds for at least one neighbor $W(x,y)$ within a $r_x \times r_y$ patch:  
\begin{equation}
    W(x,y)-W(x_o,y_o) - \lambda (\gamma\lvert x-x_o \rvert + (1-\gamma) \lvert y-y_o \rvert ) > t    
    \label{eq:heuristic_maskocc}
\end{equation}
with $\lambda,\gamma,r_x,r_y,t$ being hyper-parameters. Finally, the occluded points are warped back to obtain a mask $o$.

When a depth point is classified as occluded, we can neglect projection on both reference and target images (``NO" projection strategy). This avoids projecting the same pattern on the foreground (in the target image) and background (in the reference), which would increase ambiguity at occlusions -- ``BKGD" projection strategy.
Nonetheless, we follow a third strategy: we avoid projection and instead replace the original content in $(x,y)$ on the reference image with the content at $(x',y)$ in the target image. 
This does not alter the appearance of the correct match on foreground (rays originating from Q), yet stimulates the stereo matcher to establish a second correspondence with the same point $(x',y)$ in the target image, i.e., with pixel $(x,y)$ originating from P. 
This strategy will be referred to as ``FGD" projection.       
Additionally, points in the left border of the reference image would project patterns outside the target. Although irrelevant for traditional algorithms, we still project there to avoid artifacts in the predictions by deep stereo networks.
Fig. \ref{fig:pattern_qualitative1} shows an example of hallucinated pair.

\section{Experimental Results}

We describe our experimental setup, including implementation details, datasets, and results analysis.

\begin{table*}
\centering
\renewcommand{\tabcolsep}{12pt}
\scalebox{0.56}{
\begin{tabular}{|l|cc|rrrr|r|rrrr|r|rrrr|r|}

\multicolumn{3}{c}{} & \multicolumn{5}{c}{Midd-14} & \multicolumn{5}{c}{Midd-21} & \multicolumn{5}{c}{ETH3D} \\ 
\hline
Model & \multicolumn{2}{c|}{Depth Points} & \multicolumn{4}{c|}{Error Rate (\%)} & avg. & \multicolumn{4}{c|}{Error Rate (\%)} & avg. & \multicolumn{4}{c|}{Error Rate (\%)} & avg. \\
 & Train & Test & $>1$ & $>2$ & $>3$ & $>4$ & (px) & $>1$ & $>2$ & $>3$ & $>4$ & (px) & $>1$ & $>2$ & $>3$ & $>4$ & (px) \\
 \hline\hline
rSGM \cite{spangenberg2014large} & \xmark & \xmark 
& 69.97 & 41.86 & 29.41 & 25.19 & 13.62
& 67.33 & 40.55 & 27.69 & 22.37 & 8.02
& 27.20 & 9.11 & 5.64 & 4.26 & 1.18 \\
\hline
rSGM-{\em gd} \cite{poggi2019guided} & \xmark & \cmark 
& 63.47 & 30.64 & 17.03 & 13.57 & 9.98  
& 59.32 & 26.26 & 13.51 & 10.04 & 4.44 
& 19.95 & 3.18 & 1.72 & 1.28 & 0.76 \\
rSGM-{\em vpp} & \xmark & \cmark 
& \bf 57.98 & \bf 20.37 & \bf 10.86 & \bf 9.49 & \bf 7.81
& \bf 52.81 & \bf 15.08 & \bf 6.63 & \bf 5.51 & \bf 3.40
& \bf  9.66 & \bf 0.62 & \bf 0.46 &  \bf0.41 & \bf 0.58\\
\hline\hline

PSMNet \cite{chang2018psmnet} & \xmark & \xmark 
& 48.52 & 31.11 & 24.29 & 20.58 & 10.05 
& 47.25 & 28.03 & 20.19 & 15.88 & 4.52 
& 19.73 & 6.48 & 4.09 & 3.11 & 0.88 \\
\hline
PSMNet-{\em gd} \cite{poggi2019guided} & \xmark & \cmark 
& 48.24 & 30.66 & 24.03 & 20.47 & 10.50 
& 46.61 & 27.18 & 19.60 & 15.59 & 4.49 
& 18.96 & 5.91 & 3.56 & 2.85 & 0.84 \\
PSMNet-{\em vpp} & \xmark & \cmark 
& \bf 26.30 & \bf 16.08 & \bf 13.06 & \bf 11.60 & \bf 6.11
& \bf 22.26 & \bf 10.52 & \bf 6.96 & \bf 5.36 & \bf 1.63
& \bf 10.83 & \bf 2.01 & \bf 1.46 & \bf 1.30 & \bf 0.53 \\
\hline
PSMNet-{\em gd}-{\em ft} \cite{poggi2019guided} & \cmark & \cmark 
& 33.82 & 15.61 & 10.74 & 8.77 & 4.52 
& 35.76 & 13.77 & 7.55 & 5.21 & 1.71 
& 12.28 & 2.43 & 0.86 & 0.61 & 0.54 \\
PSMNet-{\em vpp}-{\em ft} & \cmark & \cmark 
& \bf 25.07 & \bf 14.92 & \bf 11.87 & \bf 10.43 & \bf 6.19
& \bf 21.21 & \bf 10.17 & \bf 6.78 & \bf 5.26 & \bf 1.66
& \bf 2.86 & \bf 1.40 & \bf 1.21 & \bf 1.11 & \bf 0.38\\
\hline
PSMNet-{\em gd}-{\em tr} \cite{poggi2019guided} & \cmark & \cmark 
& 25.17 & \bf 12.61 & \bf 9.13 & \bf 7.59 & \bf 3.84 
& 23.67 & 9.63 & 5.75 & \bf 4.15 & \bf 1.33 
& 4.79 & \bf 0.85 & \bf 0.54 & \bf 0.42 & \bf 0.33 \\
PSMNet-{\em vpp}-{\em tr} & \cmark & \cmark 
& \bf 21.32 & 12.95 & 10.52 & 9.35 & 5.09
& \bf 18.24 & \bf 8.57 & \bf 5.73 & 4.50 & 1.57
& \bf 2.18 & 1.48 & 1.36 & 1.29 & 0.34\\
LidarStereoNet \cite{LIDARSTEREONET} & \cmark & \cmark 
& 32.72 & 16.30 & 12.10 & 10.34 & 4.48 
& 27.32 & 11.67 & 7.58 & 5.88 & 1.80 
& 10.39 & 1.05 & 0.47 & 0.30 & 0.45 \\ 
\hline\hline
CCVNorm \cite{wang20193d} & \cmark & \cmark 
& 30.22 & 12.49 & 7.54 & 5.58 & 2.27
& 20.88 & 7.63 & 4.47 & 3.28 & 1.14
& 17.63 & 4.69 & 2.16 & 1.28 & 0.66 \\ 
\hline\hline
RAFT-Stereo \cite{lipson2021raft} & \xmark & \xmark 
& 24.24 & 15.65 & 12.48 & 10.62 & 3.87 
& 20.05 & 10.28 & 7.18 & 5.55 & 1.31 
& 2.84 & 1.44 & 0.90 & 0.74 & 0.28 \\
\hline
RAFT-Stereo-{\em gd} \cite{poggi2019guided} & \xmark & \cmark 
& 24.14 & 11.58 & 7.85 & 6.08 & 2.87  
& 20.38 & 8.31 & 5.09 & 3.65 & 1.11 
& 5.32 & 1.68 & 1.06 & 0.76 & 0.41 \\
RAFT-Stereo-{\em vpp} & \xmark & \cmark 
& \bf 8.04 & \bf 5.30 & \bf 4.35 & \bf 3.81 & \bf 2.01
& \bf 6.94 & \bf 4.26 & \bf 3.33 & \bf 2.78 & \bf 0.72
& \bf 1.30 & \bf 0.83 & \bf 0.65 & \bf 0.54 & \bf 0.15\\
\hline
RAFT-Stereo-{\em gd}-{\em ft} \cite{poggi2019guided} & \cmark & \cmark 
& 15.22 & 8.02 & 5.97 & 5.05 & 2.67
& 15.51 & 6.76 & 4.63 & 3.63 & 1.21 
& 2.52 & 1.28 & 1.00 & 0.77 & 0.29 \\
RAFT-Stereo-{\em vpp}-{\em ft} & \cmark & \cmark 
& \bf 7.12 & \bf 4.73 & \bf 3.95 & \bf 3.51 & \bf 1.95 
& \bf 6.40 & \bf 4.00 & \bf 3.20 & \bf 2.76 & \bf 0.83
& \bf 1.01 & \bf 0.74 & \bf 0.64 & \bf 0.58 & \bf 0.14\\
\hline
RAFT-Stereo-{\em gd}-{\em tr} \cite{poggi2019guided} & \cmark & \cmark 
& 6.39 & \bf 3.29 & \bf 2.35 & \bf 1.93 & \bf 0.91 
& 6.45 & 3.14 & \bf 2.25 & \bf 1.82 & \bf 0.64 
& 0.94 & 0.59 & \bf 0.46 & \bf 0.38 & 0.14 \\
RAFT-Stereo-{\em vpp}-{\em tr} & \cmark & \cmark 
& \bf 5.57 & 4.14 & 3.63 & 3.33 & 1.51
& \bf 4.85 & \bf 3.05 & 2.39 & 2.01 & 0.66
& \bf 0.71 & \bf 0.57 & 0.51 & 0.47 & \bf 0.12\\
\hline
\end{tabular}}
\vspace{-0.3cm}
\caption{\textbf{Comparison with existing methods.} Results on Midd-14, Midd-21, ETH3D. Networks trained on synthetic data.
}
\label{tab:Results_Middlebury_ETH}
\end{table*}

\subsection{Implementation and Experimental Settings}

All virtual pattern variants depicted in Fig. \ref{fig:Duck_pattern} from (ii) to (vii) are implemented in Cython, sub-pixel disparities splatted on adjacent pixels in the right view. Among the hyper-parameters, we set $\lambda=2,\gamma=0.4375,t=1$ and $r_x \times r_y = 9 \times 7$ for the occlusion detection heuristic, whose effectiveness is studied in the supplementary material.
To assess the effectiveness of VPP, we run several experiments in comparison with existing approaches that combine sparse depth points with stereo algorithms and networks. In particular, we consider the Guided Stereo Matching framework \cite{poggi2019guided}, LidarStereoNet \cite{LIDARSTEREONET}, and CCVNorm \cite{wang20193d} as main competitors. 

Since the first two approaches are implemented over the PSMNet \cite{chang2018psmnet} architecture, we apply VPP to PSMNet for a direct and fair comparison. However, the original PSMNet weights used in \cite{poggi2019guided} yielded poor generalization results; therefore, we retrain it following the original protocol \cite{chang2018psmnet}, i.e., for 10 epochs on SceneFlow with a constant learning rate equal to 1e-3. For fairness, LidarStereoNet and CCVNorm have been retrained using the same setting.
Moreover, since guided stereo \cite{poggi2019guided} represents our closest competitor -- i.e., it can be applied without any architectural change to the stereo model -- we implement it within a more modern network, RAFT-Stereo \cite{lipson2021raft}, and compare it against our VPP. 
To conclude, we extend this comparison with an implementation of the Semi-Global-Matching method \cite{hirschmuller2007stereo}, i.e., rSGM \cite{spangenberg2014large}, as a representative of traditional stereo algorithms. We run it by setting the maximum disparity to 192 disparity P1=11, adaptive P2 \cite{banz2012evaluation} (with P2$_\text{min}$=17, P2$_\alpha=0.5$, P2$_\gamma=35$), applying left-right check and a speckle filter to remove outliers, then filling holes with background interpolation \cite{Menze2015CVPR}.

\subsection{Evaluation Datasets \& Protocol}

We run experiments on three indoor/outdoor datasets.

\textbf{Middlebury}. The Middlebury dataset \cite{scharstein2014high} is a high-resolution stereo dataset featuring indoor scenes, captured under controlled lighting conditions with accurate ground-truth obtained using structured light. We evaluate our approach on three different splits: the \textit{Additional} 13 scenes in Middlebury 2014 (Midd-A), the 15 scenes from Middlebury 2014 training set (Midd-14), 
and the 24 scenes from Middlebury 2021 (Midd-21). Results are evaluated at full resolution, with PSMNet models running at half and CCVNorm at quarter (Midd-14) and one third (Midd-21) resolution.

\textbf{KITTI 2015 \cite{Menze2015CVPR}.} This real-world stereo dataset depicts autonomous driving scenarios, captured at a resolution of approximately 1280$\times$384 pixels with sparse ground-truth depth maps collected using a LiDAR sensor. It provides 200 stereo pairs annotated with ground-truth, including independently moving objects such as cars. 
Among the 200 samples, we select 142 stereo pairs for which raw LiDAR measurements are provided \cite{LIDARSTEREONET}, allowing for evaluating VPP and competitors with data from a real sensor.

\textbf{ETH3D \cite{schoeps2017cvpr}.} This dataset collects indoor and outdoor scenes, with a total of 27 grayscale low-resolution stereo pairs and corresponding ground-truth disparity maps.

\textbf{Evaluation Protocol.} We compute the percentage of pixels with a disparity error higher than a certain threshold $\tau$, with respect to the ground-truth. 
We report error rates by varying $\tau$ to 1, 2, 3, and 4, together with the average disparity error (\textit{avg}). 
We evaluate our computed disparity maps across both occluded and non-occluded regions with valid ground truth disparity unless otherwise noted.

\subsection{Ablation Study}

We start by studying the different pattern variants and components under different settings.
Tab. \ref{tab:ablation} summarizes this ablation study, 
conducted on Midd-A at full resolution with PSMNet \cite{chang2018psmnet}, RAFT-Stereo \cite{lipson2021raft} and rSGM.

For each stereo pair, we randomly sample 5\% sparse depth points from the dense ground-truth. 
Simply enabling virtual projection with pointwise virtual patterns (ii) and (iii), without explicit handling of occlusions, gives a massive boost in performance compared to the baseline, dropping the error rate to half for RAFT-Stereo (11 to 5\%) and PSMNet (29 to 15\%), while also reducing the error rate from 34 to 20\% for rSGM. 
Adding alpha-blending with pointwise patterns yields similar, yet slightly worse results across all methods. Nonetheless, projecting patterns on $3 \times 3$ patches in conjunction with alpha-blending enables additional and consistent improvements to all methods with virtual pattern (iv) the most effective except for rSGM, which is not surprisingly more effective with pattern (vii) since it builds the cost volume based on a pointwise cost function. By handling occlusions thanks to our heuristic, we obtain some further improvements when neglecting projection (``NO" strategy).
Finally, enabling occlusion handling through the ``FGD" strategy yields the best results for all methods, with slight differences among them with the four patch-based virtual patterns. Considering the outcome of the ablation, despite the slightly worse accuracy, we conduct the remaining experiments with virtual pattern vi) on $3 \times 3$ patches, because of its a negligible computation overhead with respect to histogram-based projection, while enabling at the same time alpha-blending and ``FGD" occlusion handling -- i.e., in yellow in the table.

\subsection{Comparison with Existing Approaches}

We now assess the performance of VPP and its main competitors. 
If not differently specified, we randomly sample 5\% depth points from ground-truth, as in \cite{poggi2019guided}. 

\textbf{Guided/VPP variants.} Given the flexibility of both VPP and guided stereo, we evaluate their application to stereo networks under three different settings, respectively:

\begin{itemize}
    \item Without retraining the stereo network (\textbf{-gd/-vpp})
    \item After a brief fine-tuning of the pre-trained model, enhanced by guided or VPP frameworks (\textbf{-gd-ft}/\textbf{-vpp-ft})
    \item By training from scratch a new model, enhanced by guided or VPP (\textbf{-gd-tr}/\textbf{-vpp-tr})
\end{itemize}
For \textit{-ft} variants, one epoch of finetuning is carried out on FlyingThings, with learning rate 1e-4 and 1e-5 for PSMNet and RAFT-Stereo, respectively.
For \textit{-tr} variants, PSMNet and RAFT-Stereo are trained for 10 and 20 epochs, with learning rates 1e-3 and 1e-4, respectively. In any case, PSMNet and RAFT-Stereo process $384\times512$ and $360\times720$ crops, respectively, with batch size 2 on a single 3090 GPU. 

The three variants allow for evaluating guided stereo and VPP when deployed with the least effort -- i.e., by simply taking a pre-trained stereo network off the shelf -- or with deeper intervention by the developer, either through a short fine-tuning on the pre-existing model or, with major efforts, by retraining it from scratch.

\begin{figure}
    \centering
    \includegraphics[width=1.0\linewidth]{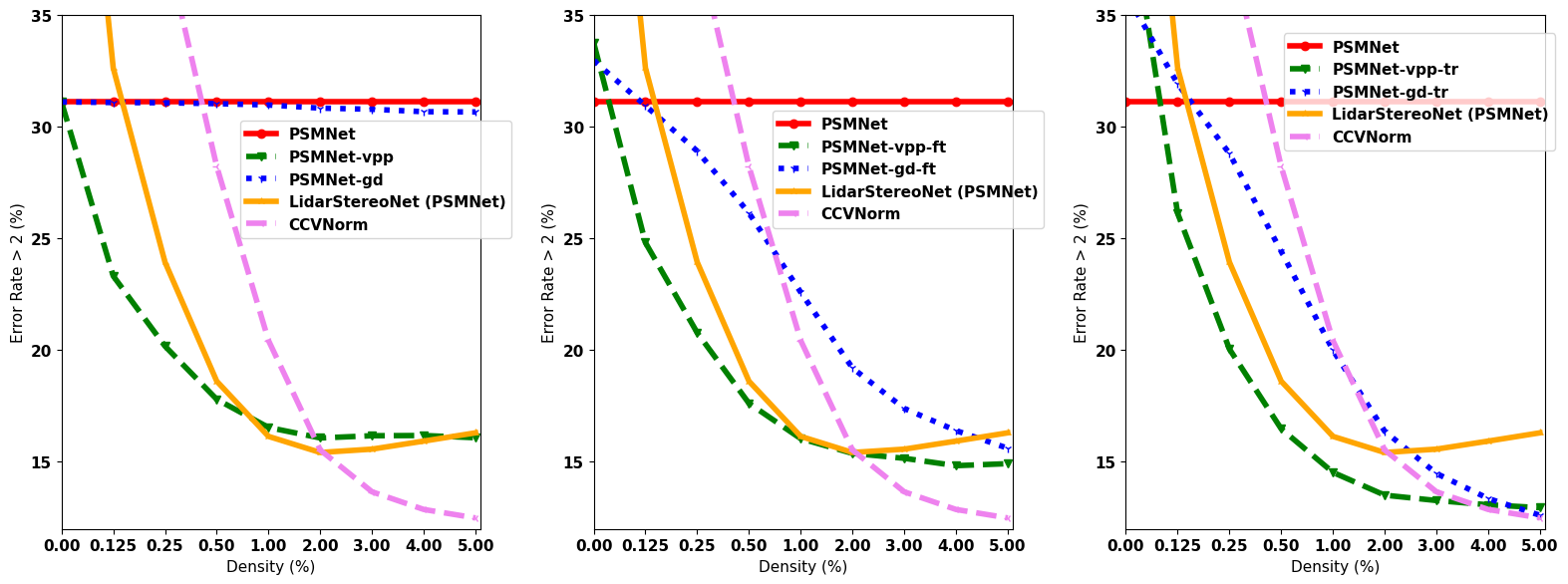}
    \includegraphics[width=1.0\linewidth]{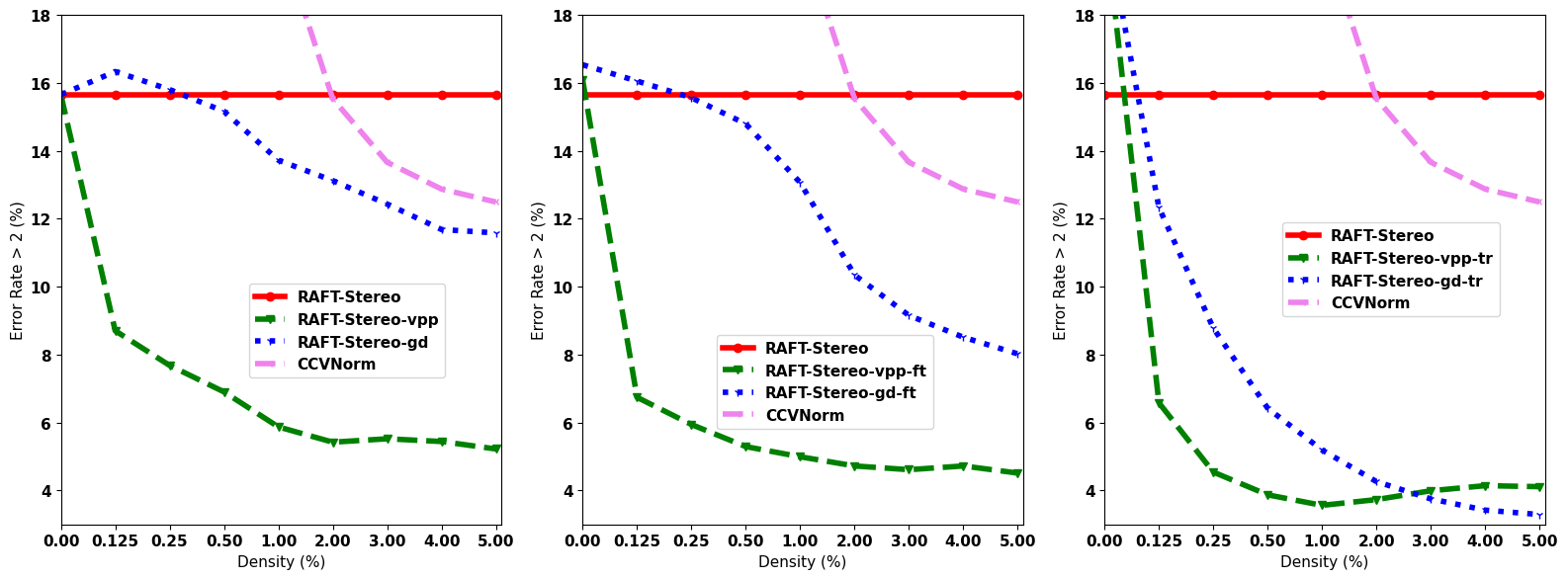}
    \vspace{-0.3cm}
    \caption{\textbf{Depth sparsity vs accuracy.} Results by VPP and competitors with different amounts of depth points.}
    \label{fig:raft_density_curve}\vspace{-0.4cm}
\end{figure}

\begin{table}
\centering
\scalebox{0.7}{
\begin{tabular}{|l|cc|rrrr|r|}

\multicolumn{3}{c}{} & \multicolumn{5}{c}{Midd-14} \\ 
\hline
Model & \multicolumn{2}{c|}{Depth Points} & \multicolumn{4}{c|}{Error Rate (\%)} & avg. \\
 & Train & Test & $>1$ & $>2$ & $>3$ & $>4$ & (px) \\
 \hline\hline
rSGM \cite{spangenberg2014large} & \xmark & \xmark &
69.97 & 41.86 & 29.41 & 25.19 & 13.62 \\
rSGM-{\em gd} & \xmark & \cmark &
63.47 & 30.64 & 17.03 & 13.57 & 9.98 \\

rSGM-{\em vpp} & \xmark & \cmark & 
57.98 & 20.37 & 10.86 & 9.49 & \bf 7.81\\
rSGM-{\em gd}-{\em vpp} & \xmark & \cmark & 
\bf 57.74 & \bf 19.55 & \bf 10.05 & \bf 8.69 & 8.62 \\
\hline\hline
PSMNet \cite{chang2018psmnet} & \xmark & \xmark & 48.52 & 31.11 & 24.29 & 20.58 & 10.05 \\
PSMNet-{\em gd} & \xmark & \cmark & 48.24 & 30.66 & 24.03 & 20.47 & 10.50 \\

PSMNet-{\em vpp} & \xmark & \cmark & 
\bf 26.30 & \bf 16.08 & \bf 13.06 & \bf 11.60 & \bf 6.11 \\
PSMNet-{\em gd}-{\em vpp} & \xmark & \cmark & 
27.68 & 16.49 & 13.44 & 12.00 & 6.30  \\

\hline\hline
LidarStereoNet \cite{LIDARSTEREONET} & \cmark & \cmark & 32.72 & 16.30 & 12.10 & 10.34 & 4.48 \\
LidarStereoNet{\em -vpp} & \cmark & \cmark & \bf 30.02 & \bf 15.32 & \bf 11.48 & \bf 9.86 & \bf 4.46 \\
\hline\hline
CCVNorm \cite{wang20193d} & \cmark & \cmark & 30.22 & 12.49 & 7.54 & 5.58 & 2.27 \\
CCVNorm-vpp & \cmark & \cmark & \bf 28.29 & \bf 12.26 & \bf 7.42 & \bf 5.51 & \bf 2.20 \\
\hline\hline
RAFT-Stereo \cite{lipson2021raft} & \xmark & \xmark & 24.24 & 15.65 & 12.48 & 10.62 & 3.87 \\
RAFT-Stereo-{\em gd} & \xmark & \cmark & 24.14 & 11.58 & 7.85 & 6.08 & 2.87 \\

RAFT-Stereo-{\em vpp} & \xmark & \cmark & 
\bf 8.04 & \bf 5.30 & \bf 4.35 & \bf 3.81 & \bf 2.01 \\
RAFT-Stereo-{\em gd}-{\em vpp} & \xmark & \cmark & 
12.85 & 6.52 & 4.78 & 3.93 & 2.10 \\
\hline
\end{tabular}}
\vspace{-0.3cm}
\caption{\textbf{Combining VPP with existing methods.}
Results on Midd-14, networks trained on synthetic data.
}
\label{tab:Results_all_together}
\end{table}

\textbf{Comparison on Middlebury/ETH3D.}
Tab. \ref{tab:Results_Middlebury_ETH} reports the outcome on Midd-14, Midd-21 and ETH3D datasets using models trained on synthetic data only or the rSGM algorithm. 
We can immediately notice one of the most prominent figures of merit of our proposal, i.e., the remarkable ability to largely boost cross-domain generalization on all datasets without any retraining of the pre-existing model (-vpp). On the contrary, guided stereo improves the results only marginally under this setting (-gd), while CCVNorm and LidarStereoNet need to be trained from scratch to process depth points (i.e., they are concatenated to RGB images). Accordingly, VPP is the undisputed winner in boosting the accuracy of stereo networks taken off the shelf.

By shortly fine-tuning the original networks to take advantage of sparse depth data is beneficial for both guided stereo (-gd-ft) and VPP (-vpp-ft) although the latter frequently, and always with RAFTStereo, outperforms the former even without any fine-tuning (-vpp), while training from scratch the networks to exploit the depth data with guided stereo (-gd-tr) results, sometimes, negligibly better than with VPP (-vpp-tr). This fact emphasizes further how VPP is much less training dependent, which is remarkable. Moreover, acting at the image level (VPP) is more robust than concatenating depth data to RGB (LidarStereoNet), acting on cost volumes (guided stereo) or both (CCVNorm).

\begin{table}
\centering
\scalebox{0.7}{
\begin{tabular}{|l|cc|rrrr|r|}

\multicolumn{3}{c}{} & \multicolumn{5}{c}{Midd-21} \\ 
\hline
Model & \multicolumn{2}{c|}{Depth Points} & \multicolumn{4}{c|}{Error Rate (\%)} & avg. \\
 & Train & Test & $>1$ & $>2$ & $>3$ & $>4$ & (px) \\
\hline\hline
PSMNet \cite{chang2018psmnet} & \xmark & \xmark & 44.75 & 24.98 & 17.31 & 13.37 & 3.42 \\
\hline
PSMNet-{\em gd} & \xmark & \cmark & 44.40 & 24.58 & 16.98 & 13.09 & 3.40 \\
PSMNet-{\em vpp} & \xmark & \cmark & 
\bf 21.38 & \bf 10.32 & \bf 6.85 & \bf 5.24 & \bf 1.51\\
\hline
PSMNet-{\em gd}-{\em ft} & \cmark & \cmark & 40.47 & 16.06 & 8.56 & 5.83 & 1.78 \\
PSMNet-{\em vpp}-{\em ft} & \cmark & \cmark & 
\bf 21.36 & \bf 10.12 & \bf 6.68 & \bf 5.12 & \bf 1.56 \\
\hline
PSMNet-{\em gd}-{\em tr} & \cmark & \cmark & 23.15 & 9.44 & \bf 5.60 & \bf 4.04 & \bf 1.29 \\
PSMNet-{\em vpp}-{\em tr} & \cmark & \cmark & 
\bf 18.07 & \bf 8.64 & 5.79 & 4.51 & 1.50 \\
LidarStereoNet \cite{LIDARSTEREONET} & \cmark & \cmark & 25.08 & 10.09 & 6.47 & 4.94 & 1.86 \\ 
\hline\hline
CCVNorm \cite{wang20193d} & \cmark & \cmark & 20.54 & 7.45 & 4.31 & 3.12 & 1.06 \\
\hline\hline
RAFT-Stereo \cite{lipson2021raft} & \xmark & \xmark & 
19.22 & 9.38 & 6.28 & 4.68 & 1.26 \\
\hline
RAFT-Stereo-{\em gd} & \xmark & \cmark & 
18.82 & 6.95 & 4.06 & 2.88 & 1.05 \\
RAFT-Stereo-{\em vpp} & \xmark & \cmark & 
\bf 6.65 & \bf 4.00 & \bf 3.10 & \bf 2.60 & \bf 0.71 \\
\hline
RAFT-Stereo-{\em gd}-{\em ft} & \cmark & \cmark & 14.87 & 6.09 & 3.99 & 3.09 & 1.02 \\
RAFT-Stereo-{\em vpp}-{\em ft} & \cmark & \cmark & 
\bf 6.68 & \bf 4.16 & \bf 3.34 & \bf 2.85 & \bf 0.82 \\
\hline
RAFT-Stereo-{\em gd}-{\em tr} & \cmark & \cmark & 6.04 & \bf 2.91 & \bf 2.08 & \bf 1.70 & \bf 0.58 \\
RAFT-Stereo-{\em vpp}-{\em tr} & \cmark & \cmark & 
\bf 4.76 & 3.00 & 2.39 & 2.02 & 0.69 \\
\hline
\end{tabular}}
\vspace{-0.3cm}
\caption{\textbf{Results with fine-tuned models.}
Results on Midd-21, after fine-tuning on Midd-14.
}\vspace{-0.5cm}
\label{tab:Results_Middlebury_2021}
\end{table}

\begin{table}
\centering
\scalebox{0.7}{
\begin{tabular}{|l|cc|rrrr|r|}

\multicolumn{3}{c}{} & \multicolumn{5}{c}{KITTI 142} \\ 
\hline
Model & \multicolumn{2}{c|}{Depth Points} & \multicolumn{4}{c|}{Error Rate (\%)} & avg. \\
 & Train & Test & $>1$ & $>2$ & $>3$ & $>4$ & (px) \\

\hline\hline
rSGM \cite{spangenberg2014large} & \xmark & \xmark 
& 43.53 & 15.66 & 8.27 & 5.68 & 1.56 \\
\hline
rSGM-{\em gd} & \xmark & \cmark 
& 35.65 & 10.00 & 5.50 & 4.03 & 1.30 \\
rSGM-{\em vpp} & \xmark & \cmark 
& \bf 23.56 & \bf 6.12 & \bf 4.04 & \bf 3.20 & \bf 1.17 \\
\hline\hline
PSMNet \cite{chang2018psmnet} & \xmark & \xmark & 32.50 & 11.70 & 6.40 & 4.51 & 1.32 \\
\hline
PSMNet-{\em gd} & \xmark & \cmark & 32.59 & 11.94 & 6.71 & 4.81 & 1.35 \\
PSMNet-{\em vpp} & \xmark & \cmark & 
\bf 20.94 & \bf 6.85 & \bf 4.24 & \bf 3.26 & \bf 1.06 \\
\hline
PSMNet-{\em gd}-{\em ft} & \cmark & \cmark & 30.39 & 9.79 & 5.18 & 3.69 & 1.27 \\
PSMNet-{\em vpp}-{\em ft} & \cmark & \cmark & 
\bf 21.97 & \bf 6.63 & \bf 3.91 & \bf 2.91 & \bf 1.07 \\
\hline
PSMNet-{\em gd}-{\em tr} & \cmark & \cmark & 25.84 & 8.30 & 4.73 & 3.47 & 1.17\\
PSMNet-{\em vpp}-{\em tr} & \cmark & \cmark & 
\bf 17.60 & \bf 5.96 & \bf 3.86 & \bf 3.01 & \bf 1.03 \\
LidarStereoNet \cite{LIDARSTEREONET} & \cmark & \cmark & 33.01 & 10.16 & 5.13 & 3.57 & 1.33 \\ 
\hline\hline
CCVNorm \cite{wang20193d} & \cmark & \cmark & 18.98 & 6.78 & 4.50 & 3.52 & 1.17 \\
\hline\hline
RAFT-Stereo \cite{lipson2021raft} & \xmark & \xmark & 24.57 & 8.74 & 5.09 & 3.66 & 1.10 \\
\hline
RAFT-Stereo-{\em gd} & \xmark & \cmark & 32.50 & 12.61 & 7.10 & 4.95 & 1.33 \\
RAFT-Stereo-{\em vpp} & \xmark & \cmark & 
\bf 15.36 & \bf 6.33 & \bf 4.31 & \bf 3.37 & \bf 0.92 \\
\hline
RAFT-Stereo-{\em gd}-{\em ft} & \cmark & \cmark & 23.90 & 8.37 & 5.01 & 3.72 & 1.13 \\
RAFT-Stereo-{\em vpp}-{\em ft} & \cmark & \cmark & 
\bf 13.86 & \bf 5.65 & \bf 3.93 & \bf 3.12 & \bf 0.89 \\
\hline
RAFT-Stereo-{\em gd}-{\em tr} & \cmark & \cmark & 15.51 & 6.30 & 4.22 & 3.27 & 0.95 \\
RAFT-Stereo-{\em vpp}-{\em tr} & \cmark & \cmark & 
\bf 11.64 & \bf 4.78 & \bf 3.46 & \bf 2.80 & \bf 0.87 \\
\hline

\end{tabular}}
\vspace{-0.3cm}
\caption{\textbf{Experiments on KITTI.} Results on the 142 split \cite{LIDARSTEREONET} with raw LiDAR. Networks trained on synthetic data.
}
\label{tab:Results_KITTI_2015}
\end{table}

\begin{table*}
\centering
\renewcommand{\tabcolsep}{8pt}
\scalebox{0.49}{
\begin{tabular}{|l|l|cc|rrrr|r|rrrr|r|rrrr|r|rrrr|r|}

\multicolumn{4}{c}{} & \multicolumn{5}{c}{Midd-14} & \multicolumn{5}{c}{Midd-21} & \multicolumn{5}{c}{ETH3D} & \multicolumn{5}{c}{KITTI 142}\\ 
\hline
 & & \multicolumn{2}{c|}{Depth Points} & \multicolumn{4}{c|}{Error Rate (\%)} & avg. & \multicolumn{4}{c|}{Error Rate (\%)} & avg. & \multicolumn{4}{c|}{Error Rate (\%)} & avg. & \multicolumn{4}{c|}{Error Rate (\%)} & avg. \\
 Model & Model name & Train & Test & $>1$ & $>2$ & $>3$ & $>4$ & (px) & $>1$ & $>2$ & $>3$ & $>4$ & (px) & $>1$ & $>2$ & $>3$ & $>4$ & (px) & $>1$ & $>2$ & $>3$ & $>4$ & (px) \\
\hline\hline
RAFT-Stereo \cite{lipson2021raft} & Middlebury & \xmark & \xmark 
& 17.77 & 9.73 & 7.16 & 5.85 & 1.67
& 19.22 & 9.38 & 6.28 & 4.68 & 1.26
& 2.78 & 1.42 & 1.00 & 0.86 & 0.29
& 24.95 & 7.81 & 4.39 & 3.19 & 1.05\\
RAFT-Stereo{\em -vpp} & Middlebury & \xmark & \cmark 
& \bf 6.78 & \bf 4.20 & \bf 3.34 & \bf 2.86 & \bf 1.11
& \bf 6.64 & \bf 3.99 & \bf 3.11 & \bf 2.60 & \bf 0.71
& \bf 1.15 & \bf 0.75 & \bf 0.59 & \bf 0.50 & \bf 0.15
& \bf 15.54 & \bf 5.45 & \bf 3.64 & \bf 2.85 & \bf 0.87\\

\hline\hline

RAFT-Stereo \cite{lipson2021raft} & ETH3D & \xmark & \xmark 
& 24.70 & 16.25 & 13.09 & 11.35 & 4.82
& 20.25 & 10.18 & 7.09 & 5.48 & 1.39
& 2.61 & 1.26 & 0.94 & 0.76 & 0.27
& 24.42 & 8.62 & 4.99 & 3.64 & 1.09\\
RAFT-Stereo{\em -vpp} & ETH3D & \xmark & \cmark 
& \bf 8.16 & \bf 5.34 & \bf 4.34 & \bf 3.79 & \bf 2.28
& \bf 6.81 & \bf 4.08 & \bf 3.17 & \bf 2.70 &  \bf0.72
& \bf 1.24 & \bf 0.77 & \bf 0.60 & \bf 0.51 & \bf 0.14
& \bf 15.16 & \bf 6.22 & \bf 4.23 & \bf 3.31 & \bf 0.92\\

\hline\hline

GMStereo \cite{xu2022unifying} & Sceneflow & \xmark & \xmark 
& 50.13 & 29.15 & 20.19 & 15.67 & 4.10
& 44.60 & 23.04 & 15.88 & 12.44 & 2.59
& 6.43 & 2.51 & 1.67 & 1.13 & 0.39 
& 30.17 & 10.46 & 5.54 & 3.85 & 1.20\\
GMStereo{\em -vpp}$^*$  & Sceneflow & \xmark & \cmark 
& \bf 28.20 & \bf 13.34 & \bf 9.25 & \bf 7.37 & \bf 2.78
& \bf 15.19 & \bf 7.12 & \bf 5.03 & \bf 4.03 & \bf 1.17
& \bf 2.45 & \bf 1.19 & \bf 0.80 & \bf 0.60 & \bf 0.25
& \bf 23.34 & \bf 7.97 & \bf 4.84 & \bf 3.62 & \bf 1.10\\
\hline\hline

GMStereo \cite{xu2022unifying} & Mixdata & \xmark & \xmark 
& 28.50 & 12.60 & 7.88 & 5.81 & 1.51
& 21.10 & 7.34 & 4.37 & 3.16 & 0.98 
& 1.72 & \bf 0.31 & \bf 0.11 & \bf 0.07 & 0.28
& 17.00 & 4.03 & \bf 2.20 & \bf 1.46 & 0.73\\
GMStereo{\em -vpp}$^*$ & Mixdata & \xmark & \cmark 
& \bf 18.32 & \bf 7.81 & \bf 5.40 & \bf 4.23 & \bf 1.34 
& \bf 11.19 & \bf 4.57 & \bf 3.09 & \bf 2.35 & \bf 0.75
& \bf 1.13 & 0.38 & 0.21 & 0.14 & \bf 0.25
& \bf 13.71 & \bf 3.82 & 2.23 & 1.56 & \bf 0.67\\
\hline\hline

CFNet \cite{Shen_2021_CVPR} & Sceneflow & \xmark & \xmark 
& 41.71 & 28.07 & 22.53 & 19.31 & 9.07
& 38.39 & 21.77 & 15.87 & 12.63 & 10.66
& 6.05 & 3.12 & 2.30 & 1.86 & 0.56
& 25.45 & 9.34 & 5.26 & 3.83 & 1.11\\

CFNet{\em -vpp}$^*$ & Sceneflow & \xmark & \cmark 
& \bf 18.51 & \bf 12.25 & \bf 10.03 & \bf 8.85 & \bf 6.78
& \bf 14.03 & \bf 8.10 & \bf 6.18 & \bf 5.07 & \bf 1.24
& \bf 1.54 & \bf 0.98 & \bf 0.76 & \bf 0.61 & \bf 0.25
& \bf 14.23 & \bf 5.97 & \bf 4.09 & \bf 3.14 & \bf 0.87\\
\hline\hline

CFNet \cite{Shen_2021_CVPR} & Middlebury & \xmark & \xmark 
& 22.38 & 11.87 & 8.40 & 6.56 & 1.91
& 28.96 & 14.44 & 9.88 & 7.45 & 2.16
& 1.15 & \bf 0.33 & \bf 0.21 & \bf 0.17 & 0.22 
& 11.08 & 3.01 & 1.73 & \bf 1.17 & 0.59 \\

CFNet{\em -vpp}$^*$ & Middlebury & \xmark & \cmark 
& \bf 13.16 & \bf 8.12 & \bf 6.33 & \bf 5.36 & \bf 1.79
& \bf 11.83 & \bf 6.92 & \bf 5.23 & \bf 4.28 & \bf 1.00 
& \bf 0.75 & \bf 0.33 & 0.25 & 0.20 & \bf 0.17 
& \bf 8.66 & \bf 2.69 & \bf 1.67 & 1.18 & \bf 0.53\\
\hline\hline

HSMNet \cite{yin2019hierarchical} & Middlebury & \xmark & \xmark 
& 31.98 & 16.53 & 11.25 & 8.61 & 2.01 
& 35.72 & 17.47 & 11.51 & 8.63 & 2.19
& 9.43 & 2.72 & 1.54 & 1.03 & 0.52
& 31.07 & 11.75 & 6.16 & 4.00 & 1.17\\
HSMNet{\em -vpp} & Middlebury & \xmark & \cmark 
& \bf 18.46 & \bf 8.96 & \bf 6.29 & \bf 5.07 & \bf 1.81
& \bf 16.96 & \bf 7.73 & \bf 5.18 & \bf 3.97 & \bf 1.13
& \bf 6.26 & \bf 1.97 & \bf 1.17 & \bf 0.86 & \bf 0.48
& \bf 27.09 & \bf 9.98 & \bf 5.47 & \bf 3.69 & \bf 1.09\\
\hline\hline

CREStereo \cite{li2022practical} & ETH3D & \xmark & \xmark 
& 19.80 & 9.99 & 7.01 & 5.44 & 1.51
& 21.81 & 9.86 & 6.76 & 5.38 & 1.36 
& 1.54 & \bf 0.51 & \bf 0.32 & \bf 0.22 & 0.19
& 21.97 & 7.26 & 4.21 & 3.07 & 0.98 \\

CREStereo{\em -vpp}$^*$ & ETH3D & \xmark & \cmark 
& \bf 12.84 & \bf 7.08 & \bf 5.31 & \bf 4.40 & \bf 1.42
& \bf 9.46 & \bf 5.69 & \bf 4.39 & \bf 3.64 & \bf 1.03
& \bf 1.17 & 0.69 & 0.50 & 0.39 & \bf 0.13
& \bf 14.76 & \bf 5.88 & \bf 3.90 & \bf 2.96 & \bf 0.85\\
\hline\hline

LEAStereo \cite{cheng2020hierarchical} & Sceneflow & \xmark & \xmark 
& 55.49 & 36.73 & 28.68 & 24.04 & 13.45
& 50.17 & 30.29 & 21.66 & 16.82 & 3.73
& 8.63 & 3.81 & 2.48 & 1.80 & 0.57
& 42.99 & 18.62 & 10.50 & 7.16 & 1.66\\
LEAStereo{\em -vpp} & Sceneflow & \xmark & \cmark 
& \bf 25.72 & \bf 14.57 & \bf 11.15 & \bf 9.50 & \bf 6.73
& \bf 19.03 & \bf 9.28 & \bf 6.32 & \bf 4.94 & \bf 1.38
& \bf 2.20 & \bf 1.20 & \bf 0.88 & \bf 0.71 & \bf 0.29 
& \bf 17.99 & \bf 6.81 & \bf 4.40 & \bf 3.34 & \bf 0.98\\
\hline
\hline
LEAStereo \cite{cheng2020hierarchical} & KITTI12 & \xmark & \xmark 
& 49.05 & 32.08 & 25.90 & 22.54 & 12.17
& 43.92 & 23.38 & 16.25 & 12.85 & 3.55
& 16.49 & 4.81 & 2.69 & 2.03 & 0.75
& 20.31 & 4.54 & 2.02 & 1.33 & 0.78 \\
LEAStereo{\em -vpp} & KITTI12 & \xmark & \cmark 
& \bf 29.17 & \bf 15.36 & \bf 12.52 & \bf 11.31 & \bf 8.11
& \bf 22.01 & \bf 10.09 & \bf 7.45 & \bf 6.21 & \bf 1.90
& \bf 11.89 & \bf 2.84 & \bf 1.66 & \bf 1.28 & \bf 0.51  
& \bf 11.98 & \bf 3.12 & \bf 1.81 & \bf 1.27 & \bf 0.63 \\

\hline
\end{tabular}}
\vspace{-0.3cm}
\caption{\textbf{VPP with off-the-shelf networks.} Results on Midd-14, Midd-21, ETH3D and KITTI. $^*$ uses $\alpha=0.2$ for blending.
}
\label{tab:roundtable}
\end{table*}

\begin{figure*}[t]
    \centering
    \includegraphics[width=0.8\textwidth]{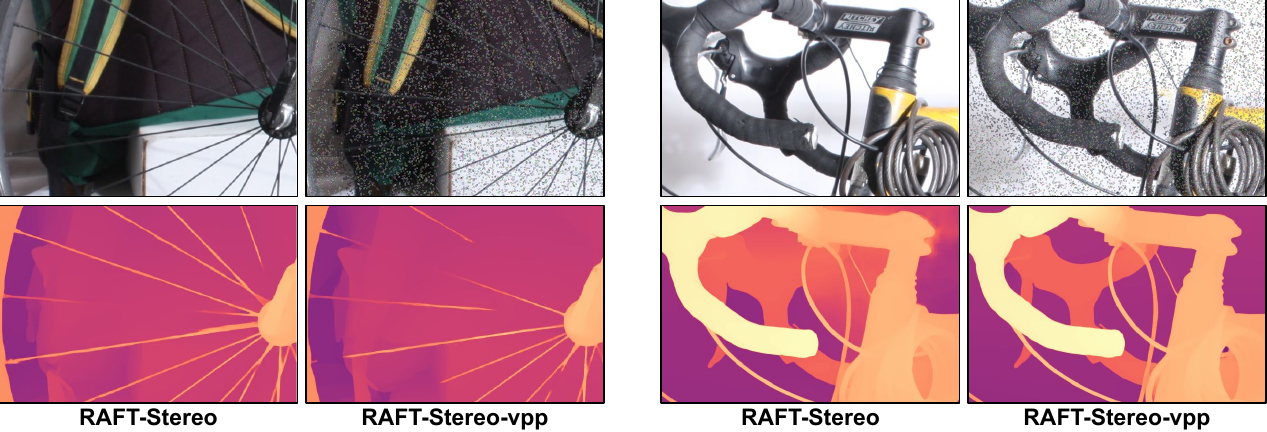}\\
    \caption{\textbf{Performance by RAFT-Stereo-vpp on thin objects.} Most fine details are preserved by VPP.}
    \label{fig:thin}
\end{figure*}

\textbf{Depth Sparsity vs Accuracy.} Fig. \ref{fig:raft_density_curve} shows the error rate ($>2$) on Midd-14 varying the density of sparse depth points from 0\% to 5\%. We can notice that VPP generally reaches almost optimal performance with a meagre 1\% density and, except few cases in the -tr configurations with some higher density, achieves much lower error rates.

\textbf{VPP + Existing Methods.} Since our approach can be used seamlessly with other image-guided methods, Tab. \ref{tab:Results_all_together} reports results yielded by such joint deployment on the Midd-14. VPP method alone is always more effective, except with rSGM, which benefits from joint deployment with guided stereo. 
Finally, using VPP with LidarStereoNet and CCVNORN also leads to slight improvements.

\textbf{Results after Fine-Tuning on Real Data.} Tab. \ref{tab:Results_Middlebury_2021} collects results analogous to those in Tab. \ref{tab:Results_Middlebury_ETH}, this time after having fine-tuned PSMNet and LidarStereoNet on Midd-14 for $\sim4000$ steps \cite{teed2020raft}, while for RAFT-Stereo we use the official Middlebury weights released by the authors. We observe a similar trend, with most VPP variants consistently outperforming the competitors, highlighting that VPP effectively improves cross-domain generalization as well as domain specialization.

 \textbf{Comparison on KITTI.} To prove VPP accuracy with noisy depth data from a real sensor outdoor and at long-range where a physical pattern would be unusable, Tab. \ref{tab:Results_KITTI_2015} reports experimental results on the KITTI 2015 dataset using raw LiDAR. 
 The results confirm the previous trends: VPP constantly outperforms guided stereo -- with any network and configuration -- LidarStereoNet and CCVNorm.

\textbf{VPP with More Off-the-shelf Networks.}
To conclude, Tab. \ref{tab:roundtable} collects the results yielded VPP applied to several off-the-shelf stereo models \cite{xu2022unifying,Shen_2021_CVPR,yin2019hierarchical,li2022practical,cheng2020hierarchical}, by running the weights provided by the authors. 
Again, VPP sensibly boosts the accuracy of any model with rare exceptions, either trained on synthetic or real data.

\subsection{Qualitative results}

To conclude, Fig. \ref{fig:thin} shows qualitatively the effect of VPP on the predictions by RAFT-Stereo. We can appreciate how our virtual pattern can greatly enhance the quality of the disparity maps, without introducing relevant artefacts in correspondence of thin structures -- despite applying the pattern on patches. 
More examples in the supplementary material.   

\section{Conclusion}

This paper proposed a novel paradigm to achieve the robustness of active stereo without the need for a pattern projector with all its inherent limitations. Purposely, our technique replaces the projector with a \textit{generic} depth sensor to obtain virtually hallucinated stereo pairs, greatly easing the visual correspondence task. Experimental results on standard stereo datasets highlight that our method achieves state-of-the-art performance with much higher flexibility, boosting the accuracy of stereo algorithms and networks.

{\small
\bibliographystyle{ieee_fullname}
\bibliography{egbib}
}

\newpage\phantom{Supplementary}
\multido{\i=1+1}{6}{


\section*{Supplementary material}

# Active Stereo Without Pattern Projector – supplementary material

Luca Bartolomei[*,†]   Matteo Poggi[†]   Fabio Tosi[†]   Andrea Conti[†]   Stefano Mattoccia[*,†]

[*]Advanced Research Center on Electronic System (ARCES)
[†]Department of Computer Science and Engineering (DISI)
University of Bologna, Italy
{luca.bartolomei5, m.poggi, fabio.tosi5, andrea.conti35, stefano.mattoccia}@unibo.it
https://vppstereo.github.io/

This document provides additional details regarding ICCV paper "Active Stereo Without Pattern Projector".

We used 5% of sparse depth points for all the experiments reported in this document.

## 1. Occlusion handling – Heuristic grid search

In this section, we report further details about our occlusion handling methodology used to estimate the occluded points in VPP. We performed a grid search to find the optimal hyper-parameters for the heuristic function used for this purpose. Specifically, we fixed the range of $r_x$ and $r_y$ to $5, 7, 9, 11, 13$, and $\lambda \in 1, 2, 3$, $\gamma \in \{0, 0.0714, 0.1429, 0.2143, 0.2857, 0.3571, 0.4286, 0.5, 0.5714, 0.6429, 0.7143, 0.7857, 0.8571, 0.9286, 1\}$, and $t \in 1, 2, 3$. To evaluate the effect of each configuration, we used the Middlebury [10] *Additional* split (Midd-A), over which we measure how good our heuristic is at classifying pixels as occluded by reporting accuracy, precision, recall and F1-score metrics. The grid search results are shown in Table I, showing only the top-50 entries sorted by F1-Score. We found that $r_x$ and $r_y$ were the most important hyper-parameters, as they remained the same for all top-50 entries. Furthermore, $\lambda$ preferred values greater than one (i.e., none of the top-50 configuration uses $\lambda = 1$), while $\gamma$ and $t$ were less influential hyper-parameters. These findings suggest that the choice of $r_x$ and $r_y$ is crucial in estimating occluded points accurately and that increasing the value of $\lambda$ can lead to improved performance.

All fifty configurations selected during the grid search show comparable performance based on F1-score. However, we chose $r_x = 9$, $r_y = 7$, $\lambda = 2$, $\gamma = 0.4375$, and $t = 1$, as they provided a slight improvement in recall. We argue that reducing false negative classification errors could be beneficial, as points classified as occluded are still projected in the "FGD"-projection configuration and thus could lead to ambiguities if not properly detected.

## 2. Complete Ablation Experiment

We extend the ablation study reported in the main paper (Tab. 1) with an exhaustive study with a larger number of configurations. Tab. II collects the results achieved by playing with the three occlusion handling strategies – i.e., "BKGD", "NO", and "FGD", respectively, in sub-tables (a), (b) and (c) –, the different virtual patterns (ii)-(vii) presented in the main paper, and also measuring the impact of Alpha-blending ($\alpha$) with different intensities

| | | | | | Midd-A | | | |
|---|---|---|---|---|---|---|---|---|
| $r_x$ | $r_y$ | $\lambda$ | $\gamma$ | $t$ | Accuracy | Precision | Recall | F1-Score |
| 9 | 7 | 3 | 0.2857 | 2 | 98.6 | 92.8 | 93.7 | 93.2 |
| 9 | 7 | 2 | 0.5714 | 3 | 98.6 | 92.7 | 93.8 | 93.2 |
| 9 | 7 | 3 | 0.5000 | 1 | 98.6 | 92.7 | 93.8 | 93.2 |
| 9 | 7 | 3 | 0.7143 | 2 | 98.6 | 92.9 | 93.6 | 93.2 |
| 9 | 7 | 3 | 0.5714 | 2 | 98.6 | 92.8 | 93.7 | 93.2 |
| 9 | 7 | 3 | 0.7857 | 3 | 98.6 | 93.0 | 93.4 | 93.2 |
| 9 | 7 | 3 | 1.0000 | 3 | 98.6 | 92.8 | 93.6 | 93.2 |
| 9 | 7 | 3 | 0.2143 | 3 | 98.6 | 93.0 | 93.5 | 93.2 |
| 9 | 7 | 2 | 0.5000 | 3 | 98.6 | 92.6 | 93.8 | 93.2 |
| 9 | 7 | 3 | 0.7143 | 1 | 98.6 | 92.6 | 93.9 | 93.2 |
| 9 | 7 | 3 | 0.7857 | 2 | 98.6 | 92.8 | 93.6 | 93.2 |
| 9 | 7 | 3 | 0.3571 | 3 | 98.6 | 93.0 | 93.5 | 93.2 |
| 9 | 7 | 3 | 0.5000 | 2 | 98.6 | 92.8 | 93.6 | 93.2 |
| 9 | 7 | 3 | 0.8571 | 2 | 98.6 | 92.8 | 93.6 | 93.2 |
| 9 | 7 | 2 | 0.3571 | 3 | 98.6 | 92.7 | 93.7 | 93.2 |
| 9 | 7 | 3 | 0.2857 | 1 | 98.6 | 92.6 | 93.8 | 93.2 |
| 9 | 7 | 3 | 0.7857 | 1 | 98.6 | 92.6 | 93.8 | 93.1 |
| 9 | 7 | 2 | 0.7857 | 2 | 98.6 | 92.5 | 93.9 | 93.1 |
| 9 | 7 | 2 | 0.6429 | 2 | 98.6 | 92.5 | 93.9 | 93.1 |
| 9 | 7 | 2 | 0.6429 | 3 | 98.6 | 92.7 | 93.7 | 93.1 |
| 9 | 7 | 3 | 0.2143 | 2 | 98.6 | 92.7 | 93.6 | 93.1 |
| 9 | 7 | 3 | 0.3571 | 1 | 98.6 | 92.6 | 93.8 | 93.1 |
| 9 | 7 | 2 | 0.8571 | 3 | 98.6 | 92.6 | 93.8 | 93.1 |
| 9 | 7 | 2 | 0.2857 | 3 | 98.6 | 92.6 | 93.8 | 93.1 |
| 9 | 7 | 3 | 0.3571 | 2 | 98.6 | 92.7 | 93.7 | 93.1 |
| 9 | 7 | 3 | 0.2143 | 1 | 98.5 | 92.5 | 93.9 | 93.1 |
| 9 | 7 | 2 | 0.7143 | 3 | 98.6 | 92.6 | 93.8 | 93.1 |
| 9 | 7 | 3 | 0.5714 | 1 | 98.6 | 92.6 | 93.7 | 93.1 |
| 9 | 7 | 3 | 0.2857 | 3 | 98.6 | 92.9 | 93.5 | 93.1 |
| 9 | 7 | 3 | 0.1429 | 2 | 98.6 | 92.6 | 93.7 | 93.1 |
| 9 | 7 | 2 | 0.3571 | 2 | 98.6 | 92.5 | 93.9 | 93.1 |
| 9 | 7 | 2 | 0.4375 | 3 | 98.5 | 92.6 | 93.8 | 93.1 |
| 9 | 7 | 2 | 0.1429 | 3 | 98.6 | 92.6 | 93.8 | 93.1 |
| 9 | 7 | 2 | 0.4375 | 2 | 98.5 | 92.4 | 94.0 | 93.1 |
| 9 | 7 | 3 | 0.6429 | 1 | 98.6 | 92.6 | 93.8 | 93.1 |
| 9 | 7 | 3 | 0.5000 | 3 | 98.6 | 93.0 | 93.3 | 93.1 |
| 9 | 7 | 2 | 0.2857 | 2 | 98.5 | 92.5 | 93.9 | 93.1 |
| 9 | 7 | 3 | 0.4375 | 2 | 98.6 | 92.7 | 93.6 | 93.1 |
| 9 | 7 | 2 | 0.5714 | 2 | 98.5 | 92.5 | 93.9 | 93.1 |
| 9 | 7 | 3 | 0.8571 | 3 | 98.6 | 92.9 | 93.5 | 93.1 |
| 9 | 7 | 3 | 0.7143 | 3 | 98.6 | 93.0 | 93.3 | 93.1 |
| 9 | 7 | 3 | 0.4375 | 1 | 98.6 | 92.6 | 93.7 | 93.1 |
| 9 | 7 | 2 | 0.5000 | 2 | 98.5 | 92.5 | 93.9 | 93.1 |
| 9 | 7 | 2 | 0.7143 | 2 | 98.5 | 92.4 | 93.9 | 93.1 |
| 9 | 7 | 3 | 0.5714 | 3 | 98.6 | 92.9 | 93.4 | 93.1 |
| 9 | 7 | 3 | 0.6429 | 3 | 98.6 | 93.0 | 93.3 | 93.1 |
| 9 | 7 | 3 | 0.6429 | 2 | 98.6 | 92.8 | 93.5 | 93.1 |
| 9 | 7 | 2 | 0.2143 | 3 | 98.5 | 92.6 | 93.7 | 93.1 |
| 9 | 7 | 2 | 1.0000 | 3 | 98.5 | 92.6 | 93.7 | 93.1 |
| 9 | 7 | 2 | 0.4375 | 1 | 98.5 | 92.3 | 94.1 | 93.1 |

Table I: **Occlusion handling – grid search.** We evaluate the effect of the hyper-parameters $r_x, r_y, \lambda, \gamma, t$.

| VPP | hyper-parameters | | | | Midd-A Error Rate (%) > 2 | | |
|---|---|---|---|---|---|---|---|
| | Pattern | α | Patch | Occ. | RAFT-St. [4] | PSMNet [1] | rSGM |
| ✗ | ✗ | ✗ | ✗ | ✗ | 11.5 | 29.3 | 34.3 |
| ✓ | (ii) | ✗ | ✗ | BKGD | 5.2 | 15.3 | 20.6 |
| ✓ | (ii) | 0.8 | ✗ | BKGD | 5.3 | 15.5 | 20.7 |
| ✓ | (ii) | 0.4 | ✗ | BKGD | 5.8 | 16.7 | 21.2 |
| ✓ | (iii) | ✗ | ✗ | BKGD | 5.1 | 15.2 | 20.2 |
| ✓ | (iii) | 0.8 | ✗ | BKGD | 5.2 | 15.3 | 20.3 |
| ✓ | (iii) | 0.4 | ✗ | BKGD | 5.6 | 16.1 | 20.5 |
| ✓ | (vi) | ✗ | 3×3 | BKGD | 5.2 | 16.0 | 15.5 |
| ✓ | (vi) | 0.8 | 3×3 | BKGD | 5.1 | 15.7 | 15.5 |
| ✓ | (vi) | 0.4 | 3×3 | BKGD | 5.0 | 15.3 | 15.9 |
| ✓ | (iv) | ✗ | 3×3 | BKGD | 4.8 | 16.0 | 16.6 |
| ✓ | (iv) | 0.8 | 3×3 | BKGD | 4.8 | 15.6 | 16.5 |
| ✓ | (iv) | 0.4 | 3×3 | BKGD | 4.9 | 15.0 | 16.7 |
| ✓ | (vii) | ✗ | 3×3 | BKGD | 4.9 | 16.1 | 16.0 |
| ✓ | (vii) | 0.8 | 3×3 | BKGD | 4.9 | 15.8 | 15.9 |
| ✓ | (vii) | 0.4 | 3×3 | BKGD | 5.0 | 15.2 | 15.9 |
| ✓ | (v) | ✗ | 3×3 | BKGD | 4.8 | 16.4 | 16.6 |
| ✓ | (v) | 0.8 | 3×3 | BKGD | 4.8 | 15.9 | 16.3 |
| ✓ | (v) | 0.4 | 3×3 | BKGD | 5.0 | 15.1 | 16.2 |
| ✓ | (vi) | ✗ | 5×5 | BKGD | 5.8 | 17.9 | 14.8 |
| ✓ | (vi) | 0.8 | 5×5 | BKGD | 5.5 | 17.1 | 14.8 |
| ✓ | (vi) | 0.4 | 5×5 | BKGD | 5.0 | 15.7 | 14.9 |
| ✓ | (iv) | ✗ | 5×5 | BKGD | 5.3 | 18.0 | 18.8 |
| ✓ | (iv) | 0.8 | 5×5 | BKGD | 5.2 | 17.3 | 17.1 |
| ✓ | (iv) | 0.4 | 5×5 | BKGD | 5.0 | 15.8 | 17.0 |
| ✓ | (vii) | ✗ | 5×5 | BKGD | 5.3 | 17.8 | 15.9 |
| ✓ | (vii) | 0.8 | 5×5 | BKGD | 5.0 | 17.2 | 15.6 |
| ✓ | (vii) | 0.4 | 5×5 | BKGD | 4.9 | 16.0 | 15.5 |
| ✓ | (v) | ✗ | 5×5 | BKGD | 5.2 | 18.1 | 19.3 |
| ✓ | (v) | 0.8 | 5×5 | BKGD | 5.1 | 17.5 | 17.3 |
| ✓ | (v) | 0.4 | 5×5 | BKGD | 4.9 | 16.3 | 16.9 |

(a)

| VPP | hyper-parameters | | | | Midd-A Error Rate (%) > 2 | | |
|---|---|---|---|---|---|---|---|
| | Pattern | α | Patch | Occ. | RAFT-St. [4] | PSMNet [1] | rSGM |
| ✗ | ✗ | ✗ | ✗ | ✗ | 11.5 | 29.3 | 34.3 |
| ✓ | (ii) | ✗ | ✗ | FGD | 5.2 | 15.0 | 20.5 |
| ✓ | (ii) | 0.8 | ✗ | FGD | 5.3 | 15.3 | 20.6 |
| ✓ | (ii) | 0.4 | ✗ | FGD | 5.8 | 16.6 | 21.1 |
| ✓ | (iii) | ✗ | ✗ | FGD | 5.2 | 14.9 | 20.1 |
| ✓ | (iii) | 0.8 | ✗ | FGD | 5.3 | 15.0 | 20.2 |
| ✓ | (iii) | 0.4 | ✗ | FGD | 5.6 | 16.0 | 20.5 |
| ✓ | (vi) | ✗ | 3×3 | FGD | 4.8 | 14.9 | 14.9 |
| ✓ | (vi) | 0.8 | 3×3 | FGD | 4.8 | 14.7 | 15.1 |
| ✓ | (vi) | 0.4 | 3×3 | FGD | 5.0 | 14.6 | 15.6 |
| ✓ | (iv) | ✗ | 3×3 | FGD | 4.6 | 14.7 | 15.7 |
| ✓ | (iv) | 0.8 | 3×3 | FGD | 4.6 | 14.6 | 15.7 |
| ✓ | (iv) | 0.4 | 3×3 | FGD | 4.8 | **14.4** | 16.1 |
| ✓ | (vii) | ✗ | 3×3 | FGD | 4.6 | 14.8 | 15.2 |
| ✓ | (vii) | 0.8 | 3×3 | FGD | 4.6 | 14.7 | 15.1 |
| ✓ | (vii) | 0.4 | 3×3 | FGD | 4.8 | **14.4** | 15.3 |
| ✓ | (v) | ✗ | 3×3 | FGD | 4.5 | 14.8 | 15.5 |
| ✓ | (v) | 0.8 | 3×3 | FGD | 4.5 | 14.6 | 15.3 |
| ✓ | (v) | 0.4 | 3×3 | FGD | 4.8 | **14.4** | 15.6 |
| ✓ | (vi) | ✗ | 5×5 | FGD | 4.6 | 15.6 | **13.6** |
| ✓ | (vi) | 0.8 | 5×5 | FGD | 4.6 | 15.2 | 13.7 |
| ✓ | (vi) | 0.4 | 5×5 | FGD | 4.7 | 14.7 | 14.0 |
| ✓ | (iv) | ✗ | 5×5 | FGD | 4.4 | 15.4 | 17.0 |
| ✓ | (iv) | 0.8 | 5×5 | FGD | **4.3** | 15.1 | 15.4 |
| ✓ | (iv) | 0.4 | 5×5 | FGD | 4.6 | 14.7 | 15.8 |
| ✓ | (vii) | ✗ | 5×5 | FGD | **4.3** | 15.6 | 14.3 |
| ✓ | (vii) | 0.8 | 5×5 | FGD | **4.3** | 15.2 | 14.1 |
| ✓ | (vii) | 0.4 | 5×5 | FGD | 4.5 | 14.6 | 14.3 |
| ✓ | (v) | ✗ | 5×5 | FGD | **4.3** | 15.5 | 17.3 |
| ✓ | (v) | 0.8 | 5×5 | FGD | **4.3** | 15.2 | 15.4 |
| ✓ | (v) | 0.4 | 5×5 | FGD | 4.4 | 14.7 | 15.4 |

(b)

| VPP | hyper-parameters | | | | Midd-A Error Rate (%) > 2 | | |
|---|---|---|---|---|---|---|---|
| | Pattern | α | Patch | Occ. | RAFT-St. [4] | PSMNet [1] | rSGM |
| ✗ | ✗ | ✗ | ✗ | ✗ | 11.5 | 29.3 | 34.3 |
| ✓ | (ii) | ✗ | ✗ | NO | 5.3 | 15.2 | 20.5 |
| ✓ | (ii) | 0.8 | ✗ | NO | 5.4 | 15.4 | 20.7 |
| ✓ | (ii) | 0.4 | ✗ | NO | 5.8 | 16.6 | 21.2 |
| ✓ | (iii) | ✗ | ✗ | NO | 5.2 | 15.0 | 20.2 |
| ✓ | (iii) | 0.8 | ✗ | NO | 5.3 | 15.2 | 20.3 |
| ✓ | (iii) | 0.4 | ✗ | NO | 5.6 | 16.0 | 20.5 |
| ✓ | (vi) | ✗ | 3×3 | NO | 5.1 | 15.5 | 15.1 |
| ✓ | (vi) | 0.8 | 3×3 | NO | 5.0 | 15.2 | 15.2 |
| ✓ | (vi) | 0.4 | 3×3 | NO | 5.1 | 14.9 | 15.7 |
| ✓ | (iv) | ✗ | 3×3 | NO | 4.8 | 15.3 | 15.9 |
| ✓ | (iv) | 0.8 | 3×3 | NO | 4.8 | 15.1 | 15.8 |
| ✓ | (iv) | 0.4 | 3×3 | NO | 4.9 | 14.7 | 16.2 |
| ✓ | (vii) | ✗ | 3×3 | NO | 4.8 | 15.5 | 15.4 |
| ✓ | (vii) | 0.8 | 3×3 | NO | 4.8 | 15.4 | 15.2 |
| ✓ | (vii) | 0.4 | 3×3 | NO | 4.9 | 14.9 | 15.3 |
| ✓ | (v) | ✗ | 3×3 | NO | 4.7 | 15.5 | 15.7 |
| ✓ | (v) | 0.8 | 3×3 | NO | 4.8 | 15.3 | 15.5 |
| ✓ | (v) | 0.4 | 3×3 | NO | 4.9 | 14.8 | 15.7 |
| ✓ | (vi) | ✗ | 5×5 | NO | 5.2 | 16.3 | 13.8 |
| ✓ | (vi) | 0.8 | 5×5 | NO | 5.1 | 15.9 | 13.8 |
| ✓ | (vi) | 0.4 | 5×5 | NO | 5.0 | 15.2 | 14.3 |
| ✓ | (iv) | ✗ | 5×5 | NO | 4.9 | 16.4 | 17.0 |
| ✓ | (iv) | 0.8 | 5×5 | NO | 4.9 | 16.0 | 15.5 |
| ✓ | (iv) | 0.4 | 5×5 | NO | 4.9 | 15.2 | 15.8 |
| ✓ | (vii) | ✗ | 5×5 | NO | 4.8 | 16.4 | 14.4 |
| ✓ | (vii) | 0.8 | 5×5 | NO | 4.8 | 16.2 | 14.2 |
| ✓ | (vii) | 0.4 | 5×5 | NO | 4.8 | 15.5 | 14.3 |
| ✓ | (v) | ✗ | 5×5 | NO | 4.8 | 16.6 | 17.4 |
| ✓ | (v) | 0.8 | 5×5 | NO | 4.7 | 16.2 | 15.5 |
| ✓ | (v) | 0.4 | 5×5 | NO | 4.8 | 15.5 | 15.6 |

(c)

Table II: **Ablation on main projection hyper-parameters.** Results on Midd-A. Networks trained on synthetic data.

(none (i.e., 1.0, 0.4 and 0.8) and the patch size (none (i.e., $1 \times 1$), $3 \times 3$ or $5 \times 5$).

As in the main paper, we study the impact of the different VPP configurations on the disparity maps predicted by RAFT-Stereo [4], PSMNet [1] or estimated by the rSGM algorithm [7] on the Middlebury [10] *Additional* split (Midd-A). For RAFT-Stereo and PSMNet, we use weights obtained after training on synthetic images only. From a first look, we can notice how determining the best overall configuration is not trivial: indeed, the configuration we select for the experiments in the main paper (in yellow) is not the absolute winner in this study, yet allowing to outperform existing techniques already [6, 2]. Nonetheless, from this exhaustive study, we draw some general take-home messages to define the best set of configurations to deploy VPP.

**Occlusion-handling.** By comparing sub-tables (a), (b) and (c), we can notice that a proper handling of the pattern projection near occlusions is helpful, most times, to attain the best results. Indeed, applying the virtual pattern according to "FGD" strategy (b) – i.e., by encouraging the matching of an occluded region with its own occluder – yields better results with respect to not handling occlusions at all (a) or to the naïve projection of the pattern in the not occluded regions only (c), when the remaining hyper-parameters remain unchanged.

**Patch size.** In general, projecting a pattern on local patches (iv)-(vii) rather than single pixels (ii)-(iii) yields better results. Not surprisingly, acting on a local region allows for easing the visual correspondence task much more than intervening on single pixels alone. Increasing the patch size from $3 \times 3$ to $5 \times 5$ can further improve the results with rSGM, while it yields mixed results with RAFT-Stereo (often improving) and PSMNet (which seems to benefit more from $3 \times 3$ patches). We argue that this behavior is caused by the higher-level cues used by deep stereo networks to perform matching, not necessarily affected by the local visual distinctiveness in the same manner as traditional algorithms.

**Alpha-blending** ($\alpha$). Since image content is crucial for modern stereo networks to compute correspondences, a proper Alpha-blending of the projected pattern with the original color images is often beneficial. In particular, it allows for compensating for the possibly erroneous correspondence being enforced by large patches near object boundaries. On the contrary, Alpha-blending always reduces the effectiveness of the virtual pattern when running rSGM, since it dampens the high distinctiveness of the pattern, which is crucial to ease correspondences when using hand-crafted matching functions.

**Different networks/algorithms.** Finally, our exhaustive ablation study highlights how different stereo networks and algorithms react differently to the configurations we experimented. Indeed, depending on the network or algorithm, we have different configurations yielding the best results. Specifically, RAFT-stereo [4] obtains the best results with $5 \times 5$ patterns (iv)-(v) or (vii), often by setting a high Alpha-blending factor ($\alpha = 0.8$) or no Alpha-blending factor at all ($\alpha = 1.0$); PSMNet [1], on the contrary, registers the best improvement with $3 \times 3$ patterns (iv)-(v) or (vii) and a moderate Alpha-blending factor ($\alpha = 0.4$); finally, rSGM [7] attains the best results with $5 \times 5$ pattern (vi), without any Alpha-blending being applied. For this reason, we selected a configuration for our main experiments that could balance the accuracy of the three methods without favoring any in particular.

| | | Depth Points | | Midd-14 | | | | | Midd-21 | | | | | ETH3D | | | | | KITTI 142 | | | | | |
| | | | | Error Rate std. (%) | | | | avg. std. | Error Rate std. (%) | | | | avg. std. | Error Rate std. (%) | | | | avg. std. | Error Rate std. (%) | | | | avg. std. |
| Model | Model name | Train | Test | >1 | >2 | >3 | >4 | (px) | >1 | >2 | >3 | >4 | (px) | >1 | >2 | >3 | >4 | (px) | >1 | >2 | >3 | >4 | (px) |
|---|---|---|---|---|---|---|---|---|---|---|---|---|---|---|---|---|---|---|---|---|---|---|---|
| RAFT-Stereo-*vpp* [4] | Sceneflow | ✗ | ✓ | 0.06 | 0.05 | 0.05 | 0.07 | 0.04 | 0.00 | 0.02 | 0.00 | 0.02 | 0.00 | 0.10 | 0.08 | 0.07 | 0.06 | 0.01 | 0.00 | 0.00 | 0.00 | 0.00 | 0.00 |
| RAFT-Stereo-*vpp* [4] | Middlebury | ✗ | ✓ | 0.06 | 0.07 | 0.09 | 0.08 | 0.02 | 0.06 | 0.07 | 0.05 | 0.06 | 0.02 | 0.07 | 0.05 | 0.04 | 0.03 | 0.00 | 0.03 | 0.00 | 0.02 | 0.01 | 0.00 |
| RAFT-Stereo-*vpp* [4] | ETH3D | ✗ | ✓ | 0.08 | 0.08 | 0.11 | 0.13 | 0.02 | 0.04 | 0.04 | 0.04 | 0.03 | 0.01 | 0.03 | 0.03 | 0.04 | 0.03 | 0.01 | 0.02 | 0.02 | 0.01 | 0.01 | 0.00 |
| GMStereo-*vpp** [11] | Sceneflow | ✗ | ✓ | 0.01 | 0.01 | 0.05 | 0.06 | 0.01 | 0.05 | 0.04 | 0.05 | 0.05 | 0.01 | 0.02 | 0.00 | 0.00 | 0.00 | 0.00 | 0.00 | 0.00 | 0.00 | 0.00 | 0.00 |
| GMStereo-*vpp** [11] | Mixdata | ✗ | ✓ | 0.00 | 0.03 | 0.02 | 0.02 | 0.00 | 0.07 | 0.03 | 0.05 | 0.06 | 0.01 | 0.02 | 0.00 | 0.00 | 0.00 | 0.00 | 0.00 | 0.00 | 0.00 | 0.00 | 0.00 |
| CFNet-*vpp** [9] | Sceneflow | ✗ | ✓ | 0.12 | 0.11 | 0.10 | 0.09 | 0.23 | 0.00 | 0.00 | 0.01 | 0.01 | 0.00 | 0.00 | 0.00 | 0.00 | 0.00 | 0.00 | 0.00 | 0.00 | 0.00 | 0.00 | 0.00 |
| CFNet-*vpp** [9] | Middlebury | ✗ | ✓ | 0.04 | 0.03 | 0.02 | 0.03 | 0.02 | 0.01 | 0.01 | 0.01 | 0.01 | 0.00 | 0.03 | 0.00 | 0.00 | 0.00 | 0.00 | 0.00 | 0.00 | 0.00 | 0.00 | 0.00 |
| HSMNet-*vpp* [12] | Middlebury | ✗ | ✓ | 0.02 | 0.00 | 0.01 | 0.02 | 0.00 | 0.06 | 0.01 | 0.01 | 0.00 | 0.01 | 0.14 | 0.02 | 0.03 | 0.02 | 0.01 | 0.00 | 0.00 | 0.00 | 0.00 | 0.00 |
| CREStereo-*vpp** [3] | ETH3D | ✗ | ✓ | 0.04 | 0.02 | 0.04 | 0.04 | 0.01 | 0.04 | 0.04 | 0.06 | 0.05 | 0.03 | 0.05 | 0.03 | 0.02 | 0.00 | 0.00 | 0.00 | 0.00 | 0.00 | 0.00 | 0.00 |
| PSMNet-*vpp* [1] | Sceneflow | ✗ | ✓ | 0.07 | 0.06 | 0.05 | 0.04 | 0.02 | 0.01 | 0.00 | 0.01 | 0.00 | 0.00 | 0.03 | 0.04 | 0.03 | 0.02 | 0.00 | 0.00 | 0.00 | 0.00 | 0.00 | 0.00 |
| rSGM-*vpp* [7] | - | ✗ | ✓ | 0.01 | 0.02 | 0.00 | 0.01 | 0.09 | 0.00 | 0.00 | 0.01 | 0.01 | 0.01 | 0.02 | 0.01 | 0.01 | 0.01 | 0.00 | 0.00 | 0.00 | 0.00 | 0.00 | 0.00 |

Table III: **Stability of random pattern of type (vi).** We present an analysis of the standard deviation of type (vi) virtual pattern based on a sample size of 5. Our findings indicate that, despite the non-deterministic nature of this pattern, the observed errors exhibit stability. * uses $\alpha = 0.2$ for blending.

## 3. Stability of random pattern

As stated in the main paper, the random patch pattern (vi) proposed and utilized is a non-deterministic technique. To assess the stability of this approach concerning artifacts, we conduct five inference runs to compute the standard deviation for various stereo methods, as many as we used in the main paper to calculate the mean errors. Our results, as shown in Table III, demonstrate that despite the non-deterministic nature of the approach, errors remain relatively constant. This evidence indicates that this pattern can be effectively employed in diverse environments and attain stable results. Specifically, the standard deviation remains below 0.14% for any error rate threshold and below 0.1 px for avg. error, except for CFNet [9] being slightly less stable (0.23 px on Midd-14).

## 4. Additional implementation details

When evaluating the accuracy of stereo algorithms and networks with and without VPP, different resizing strategies were applied to the input images, often following the suggestions of the original authors.

RAFT-Stereo [4] and HSMNet [12] were evaluated using the original image resolution on Middlebury 2014 [7], Middlebury 2021 [10], ETH3D [8], and KITTI [5] datasets. CFNet [9], rSGM [7], and PSMNet [1] were evaluated using half the original image resolution on Middlebury 2014 and Middlebury 2021 (because of memory limitations). GMStereo [11] and CREStereo [3] on Middlebury 2014 and Middlebury 2021 were evaluated with image resizing to $1024 \times 1536$. GMStereo on ETH3D was evaluated with image resizing to $512 \times 768$, while CREStereo on ETH3D was evaluated with image resizing to $768 \times 1024$. GMStereo on KITTI was evaluated with image resizing to $352 \times 1216$, while CREStereo on KITTI was evaluated with image resizing to $1024 \times 1536$.

## 5. Qualitative results

We conclude by showing some additional qualitative results to support the effectiveness of VPP further.

**Middlebury 2021.** Fig. I shows a scene from the Middlebury 2021 dataset, in which we sample 5% points from the ground-truth to project our virtual pattern. We test four stereo networks trained on synthetic data and the rSGM algorithm. We can notice how the original disparity maps suffer from several artifacts and, in most cases, very high error rates. Only RAFT-Stereo achieves reasonable results. In contrast, by enabling VPP, the results by any method fall below 13% error rate, with RAFT-Stereo achieving a bad2 error lower than 2%. To better highlight the effect of VPP on images, we show both the original and hallucinated frames in Fig. II.

**KITTI 2015.** Fig. III shows an example from the KITTI 2015 training set. Specifically, we select the *000104* stereo pair, being known by practitioners in the field for its very low exposure. Since raw LiDAR measurements are not available for this specific pair, we sample again 5% points from ground-truth. We can appreciate how, even in this prohibitive environment, VPP can largely improve the results of stereo networks and the rSGM algorithm. To better highlight the effect of VPP on images, we show both the original and hallucinated frames in Fig. IV.

| Model | RGB left | RGB right | Disparity map map Error Rate (> 2) % | VPP left | VPP right | VPP disparity map Error Rate (> 2) % |
|---|---|---|---|---|---|---|
| PSMNet [1] | | | 74.36 | | | 7.16 |
| RAFT-Stereo [4] | | | 15.91 | | | 1.94 |
| CFNet* [9] | | | 77.76 | | | 11.59 |
| GMStereo* [11] | | | 81.40 | | | 8.65 |
| rSGM [7] | | | 81.31 | | | 12.70 |

Figure I: **Cross-domain generalization enhancement: qualitative results on indoor.** All networks are trained on synthetic data and tested on an unseen scenario (Midd-21, scene podium1): our VPP approach significantly improves cross-domain generalization compared to the vanilla baselines. Additionally, traditional stereo algorithms, such as rSGM, also benefit from our approach. * uses $\alpha = 0.2$ for blending.

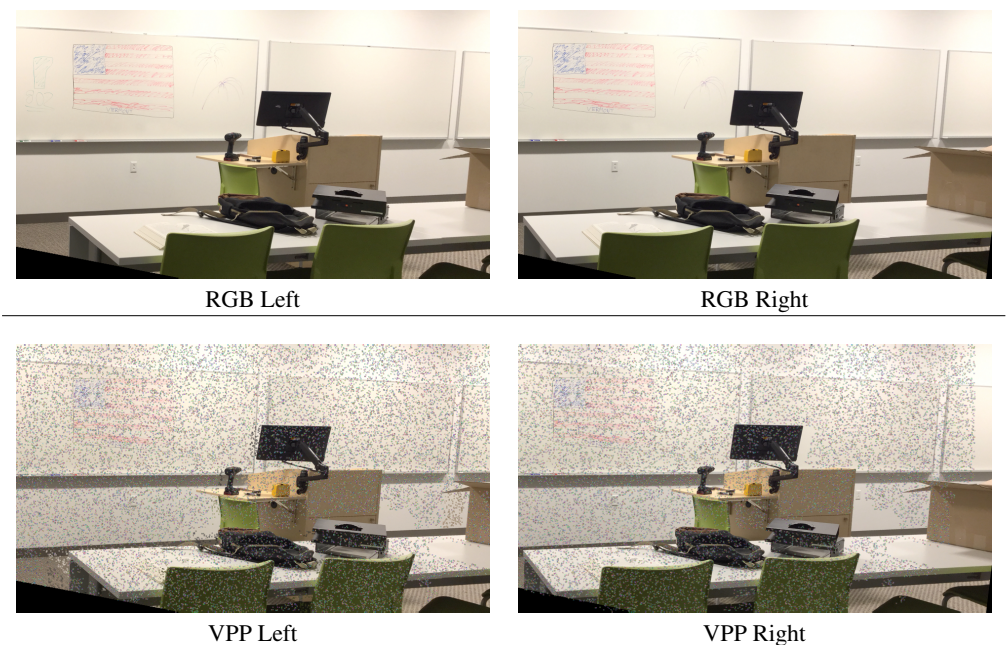

Figure II: **Qualitative view of virtual pattern in indoor scenario.** Midd-21, scene podium1

| Model | RGB left | RGB right | Disparity map map Error Rate (> 3) % | VPP left | VPP right | VPP disparity map Error Rate (> 3) % |
|---|---|---|---|---|---|---|
| PSMNet [1] | | | 69.81 | | | 1.04 |
| RAFT-Stereo [4] | | | 72.0 | | | 0.78 |
| CFNet [9] | | | 99.98 | | | 57.77 |
| GMStereo* [11] | | | 66.48 | | | 2.94 |
| rSGM [7] | | | 72.00 | | | 5.40 |

Figure III: **Cross-domain generalization enhancement: qualitative results on outdoor.** All networks are trained on synthetic data and tested on an unseen scenario (KITTI 2015, scene 104): our VPP approach significantly improves cross-domain generalization compared to the vanilla baselines, except for CFNet which achieves mild results. Additionally, traditional stereo algorithms, such as rSGM, also benefit from our approach. * uses $\alpha = 0.2$ for blending.

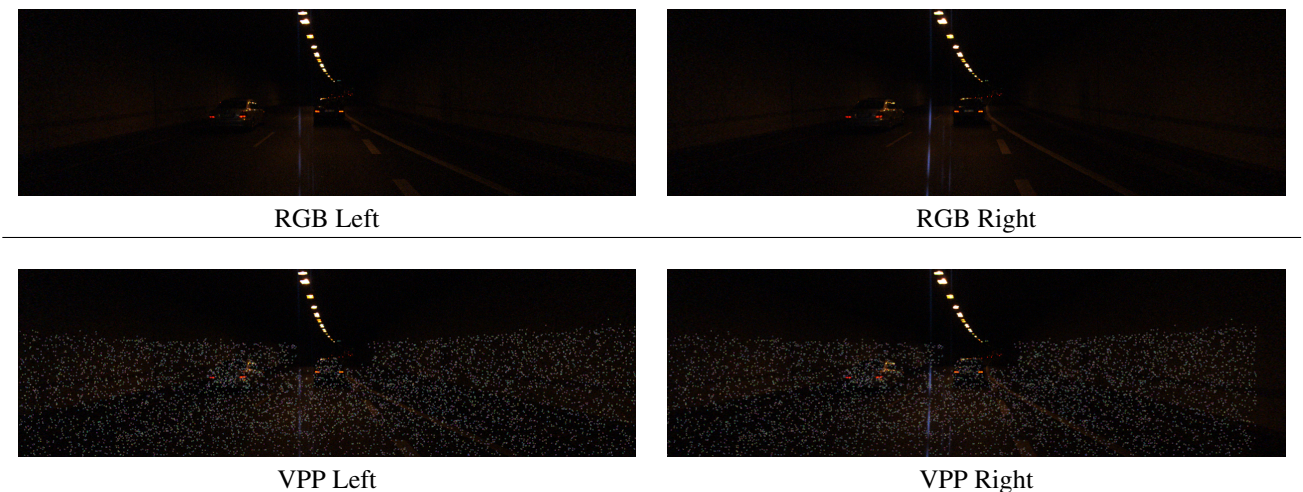

Figure IV: **Qualitative view of virtual pattern in outdoor scenario.** KITTI 2015, scene 104


}

\end{document}